\pdfoutput=1

\documentclass[11pt]{article}

\usepackage[final]{acl}

\usepackage{times}
\usepackage{latexsym}

\usepackage[T1]{fontenc}

\usepackage[utf8]{inputenc}

\usepackage{microtype}

\usepackage{inconsolata}

\usepackage{graphicx}
\usepackage{subcaption}

\usepackage{microtype}
\usepackage{hyperref}
\usepackage{url}
\usepackage{booktabs}

\usepackage{lineno}
\usepackage{graphicx}
\usepackage{makecell}
\usepackage{multirow}
\usepackage{wrapfig}
\usepackage{caption}
\usepackage{xspace}
\usepackage{amsmath}
\usepackage{enumitem}              
\usepackage{amssymb} 
\usepackage{titlesec}
\usepackage[dvipsnames]{xcolor}    
\usepackage[table]{xcolor}   
\definecolor{bluerow}{HTML}{EDF5FF}  
\definecolor{yellowrow}{HTML}{FFF8B0}
\definecolor{lightgray}{gray}{0.9} 
\definecolor{softgreen}{RGB}{230, 255, 230}

\definecolor{sagegreen}{RGB}{210, 230, 210}

\definecolor{pastelgreen}{HTML}{D1E7DD}

\usepackage{tcolorbox}
\usepackage{soul}
\tcbuselibrary{skins, breakable}

\definecolor{modelblue}{RGB}{230, 235, 255}
\definecolor{deltayellow}{RGB}{255, 255, 224}
\definecolor{textred}{RGB}{200, 0, 0}        
\definecolor{textgreen}{RGB}{0, 150, 0}      

\newcommand{\txtred}[1]{\textcolor{textred}{#1}}
\newcommand{\txtgreen}[1]{\textcolor{textgreen}{#1}}

\newcommand{\mydelta}[1]{%
  \colorbox{deltayellow}{%
    \scriptsize 
    \,+#1\, 
  }%
}

\DeclareRobustCommand{\mybasemethod}{\textsc{ThinkLogit}\xspace}
\DeclareRobustCommand{\mymethod}{\textsc{ThinkLogit-DPO}\xspace}

\newtcolorbox{tracebox}[1][]{
  colback=gray!5!white,
  colframe=gray!60!black,
  title=\textbf{\mybasemethod Long CoT Excerpt},
  fonttitle=\small,
  fontupper=\small,
  arc=0mm,
  boxrule=0.5pt,
  breakable,       
  enhanced,        
  #1
}

\newtcolorbox{guiderbox}[1][]{
  colback=gray!5!white,
  colframe=gray!60!black,
  title=\textbf{(Guider) R1-Distill-Qwen-1.5B},
  fonttitle=\small,
  fontupper=\small,
  arc=0mm,
  boxrule=0.5pt,
  breakable,       
  enhanced,        
  #1
}

\newtcolorbox{targetbox}[1][]{
  colback=gray!5!white,
  colframe=gray!60!black,
  title=\textbf{(Target) Qwen2.5-32B},
  fonttitle=\small,
  fontupper=\small,
  arc=0mm,
  boxrule=0.5pt,
  breakable,       
  enhanced,        
  #1
}

\newtcolorbox{thinklogitbox}[1][]{
  colback=gray!5!white,
  colframe=gray!60!black,
  title=\textbf{\textsc{ThinkLogit}},
  fonttitle=\small,
  fontupper=\small,
  arc=0mm,
  boxrule=0.5pt,
  breakable,       
  enhanced,        
  #1
}
\title{Logit Arithmetic Elicits Long Reasoning Capabilities Without Training}

\author{Yunxiang Zhang\thanks{$\,\,\,$Correspondence to \tt{yunxiang@umich.edu}}\hspace{3pt}\quad Muhammad Khalifa \quad Lechen Zhang\quad Xin Liu \\
\textbf{Ayoung Lee} \quad
\textbf{Xinliang Frederick Zhang} \quad \textbf{Farima Fatahi Bayat}  \quad
\textbf{Lu Wang} \\
University of Michigan
}

\begin{document}
\maketitle
\begin{abstract}
Large reasoning models exhibit long chain-of-thought reasoning with complex strategies such as backtracking and self-verification. 
Yet, these capabilities typically require resource-intensive post-training.
We investigate whether such behaviors can be elicited in large models without \textit{any} gradient updates.
To this end, we propose a decoding-time approach, \textbf{\mybasemethod}, which utilizes logit arithmetic~\citep{liutuning} to transfer these capabilities from a substantially smaller reasoning guider to a large non-reasoning target.
We further show that we can boost performance by training the guider to correct the target's errors using preference optimization over mixed model outputs, a setup we refer to as \textbf{\mymethod}. 
We evaluate these methods across six reasoning benchmarks spanning math, science, and coding domains using the Qwen2.5-32B guided by R1-Distill-Qwen-1.5B, a model 21x smaller. 
Our experiments demonstrate that \mybasemethod and \mymethod achieve a relative improvement of 21.5\% and 24.2\%, respectively, over the target model.
Moreover, \mybasemethod remains effective even when the guider and target come from different model families.  
Crucially, our method requires zero training for the large model and would incur minimal inference overhead when logits are computed in parallel, presenting a practical solution for enabling long reasoning at scale.\footnote{Our code is publicly avaiable at \url{https://github.com/yunx-z/think-logit}.}

\end{abstract}

\section{Introduction}

Large reasoning models (LRMs), such as OpenAI o1~\citep{OpenAI2024LearningToReason} and  DeepSeek-R1~\citep{DBLP:journals/corr/abs-2501-12948}, have significantly advanced reasoning by leveraging inference-time compute~\citep{DBLP:journals/corr/abs-2408-03314,DBLP:journals/corr/abs-2407-21787}. These models generate very long chain-of-thought (CoT) traces involving planning, reflection, and self-correction~\citep{DBLP:journals/corr/abs-2503-01307}. 
It is widely believed that such behaviors require specialized training, either through reinforcement learning (RL) with verifiable rewards~\citep{DBLP:journals/corr/abs-2501-12948,DBLP:journals/corr/abs-2411-15124,DBLP:journals/corr/abs-2402-03300} or supervised distillation~\citep{DBLP:journals/corr/abs-2501-19393,li2025llms} from other LRMs.
However, such training is costly for large models with long generations and many parameters. 
Meanwhile, existing \textit{training-free} long CoT elicitation methods~\citep{DBLP:journals/corr/abs-2502-03860,DBLP:journals/corr/abs-2501-19393,DBLP:journals/corr/abs-2310-01405,tang2025unlocking,zhao2025activation} still remain limited, as they often lengthen outputs without reliably inducing complex reasoning behaviors and further require domain-specific supervision or white-box access.
While the training costs are often \textit{prohibitive} for large models, small models can be trained with modest compute~\citep{dang2025reinforcement,deepscaler2025}. This observation motivates our central research question: 
\textbf{\textit{Can a small reasoning model elicit long CoT behavior in a large non-reasoning model at inference time, without training the large model?}}

We address this question with a decoding-time method, \textbf{\mybasemethod}, which elicits long CoT reasoning in a large non-reasoning model as the target. At each decoding step, we use logit arithmetic~\citep{liutuning} by computing the logit difference between a small guider model trained for long reasoning and its base version, and add the resulting shift to the target logits. This token-by-token guidance transfers long reasoning signals from the small model to the large one without requiring any gradient updates of the target.

Furthermore, since the output distributions of long and short CoTs differ substantially, we align them by training the small guider to correct errors made by the target model while maintaining the strengths of the target model.
This training process uses Direct Preference Optimization~\citep[DPO;][]{DBLP:conf/nips/RafailovSMMEF23} on mixed preference pairs sampled from both the guider and target models, thereby making \mybasemethod more \textit{on-policy}~\citep{DBLP:conf/iclr/AgarwalVZSGGB24,lu2025onpolicydistillation}, and then applies logit arithmetic using the fine-tuned guider.
We call this approach \textbf{\mymethod} and show that it further boosts performance compared to \mybasemethod.

We evaluate our methods on six challenging benchmarks covering mathematical, scientific, and code reasoning. 
Our results show that fusing the logits of a small reasoning model (R1-Distill-Qwen-1.5B) with those of a large target (Qwen2.5-32B) yields 21.5\% and 24.2\% relative performance gains with \mybasemethod and \mymethod, respectively, over the frozen target baseline.
We further demonstrate that our method generalizes effectively across model families. Notably, a Qwen-based guider successfully drives long reasoning in Llama~\citep{DBLP:journals/corr/abs-2407-21783}, OLMo~\citep{DBLP:journals/corr/abs-2501-00656}, and EXAONE~\citep{DBLP:journals/corr/abs-2412-04862} targets.
Crucially, our approach requires zero training for the large model and incurs minimal inference latency overhead by computing the guider and target logits in parallel. 

Our approach is particularly favorable when training the target is infeasible: \textit{extreme-scale} models beyond practical fine-tuning budgets~\cite{DBLP:conf/nips/DettmersPHZ23}, \textit{black-box} models limited to logit access~\cite{DBLP:conf/emnlp/OrmazabalAA23},\footnote{For example, the OpenAI API exposes a \texttt{logit\_bias} parameter that lets users alter token probabilities at inference by boosting or suppressing specific tokens~\cite{openai_logitbias}.}  or \textit{privacy-preserving} settings where an on-device guider steers a centralized model without exposing private data~\cite{DBLP:conf/aistats/McMahanMRHA17,DBLP:journals/corr/abs-2409-00088}. In these important scenarios, it enables efficient reasoning transfer without full-model fine-tuning.

To summarize, our main contributions are:
\begin{itemize}[itemsep=2pt, topsep=2pt, parsep=0pt, partopsep=0pt]
    \item We propose \textbf{\mybasemethod}, a \textit{training-free} decoding method that transfers long reasoning capabilities from a small guider to a large target using logit arithmetic (Section~\ref{sec:thinklogit}).

    \item We introduce \textbf{\mymethod}, which further boosts performance by training the guider to correct target errors via preference optimization on mixed outputs (Section~\ref{sec:thinklogit-dpo}).

    \item We demonstrate up to 24.2\% relative improvement across diverse domains (math, science, coding) and effective generalization to heterogeneous model families (Section~\ref{sec:results} \&~\ref{sec:cross_family}).
\end{itemize}

\section{Related Work}
\label{subsec:related-work}

\paragraph{Eliciting Long Chain-of-Thought Reasoning.}
Standard approaches rely on \textbf{reinforcement learning}~\citep{DBLP:journals/corr/abs-2411-15124,DBLP:journals/corr/abs-2402-03300,DBLP:journals/corr/abs-2503-14476,liu2025understanding} or \textbf{supervised fine-tuning}~\citep{DBLP:journals/corr/abs-2501-19393,DBLP:journals/corr/abs-2501-11284,DBLP:journals/corr/abs-2502-03387,li2025llms} to elicit long CoT capabilities. However, applying these to large models incurs prohibitive computational costs. 
Alternatively, \textbf{training-free} methods exploit the fact that pretrained LLMs already exhibit latent long CoT behaviors~\citep{liu2025understanding,DBLP:journals/corr/abs-2503-01307}. 
While techniques such as representation engineering~\citep{DBLP:journals/corr/abs-2310-01405,tang2025unlocking} or neuron amplification~\citep{zhao2025activation} attempt to surface these capabilities, they typically require white-box access and domain-specific supervision. 
In contrast, our \textbf{\mybasemethod} framework operates without parameter access or task-specific constraints, effectively emulating the benefits of costly training in a flexible, black-box compatible manner.

\begin{figure*}
  \centering
  \includegraphics[width=0.9\linewidth]{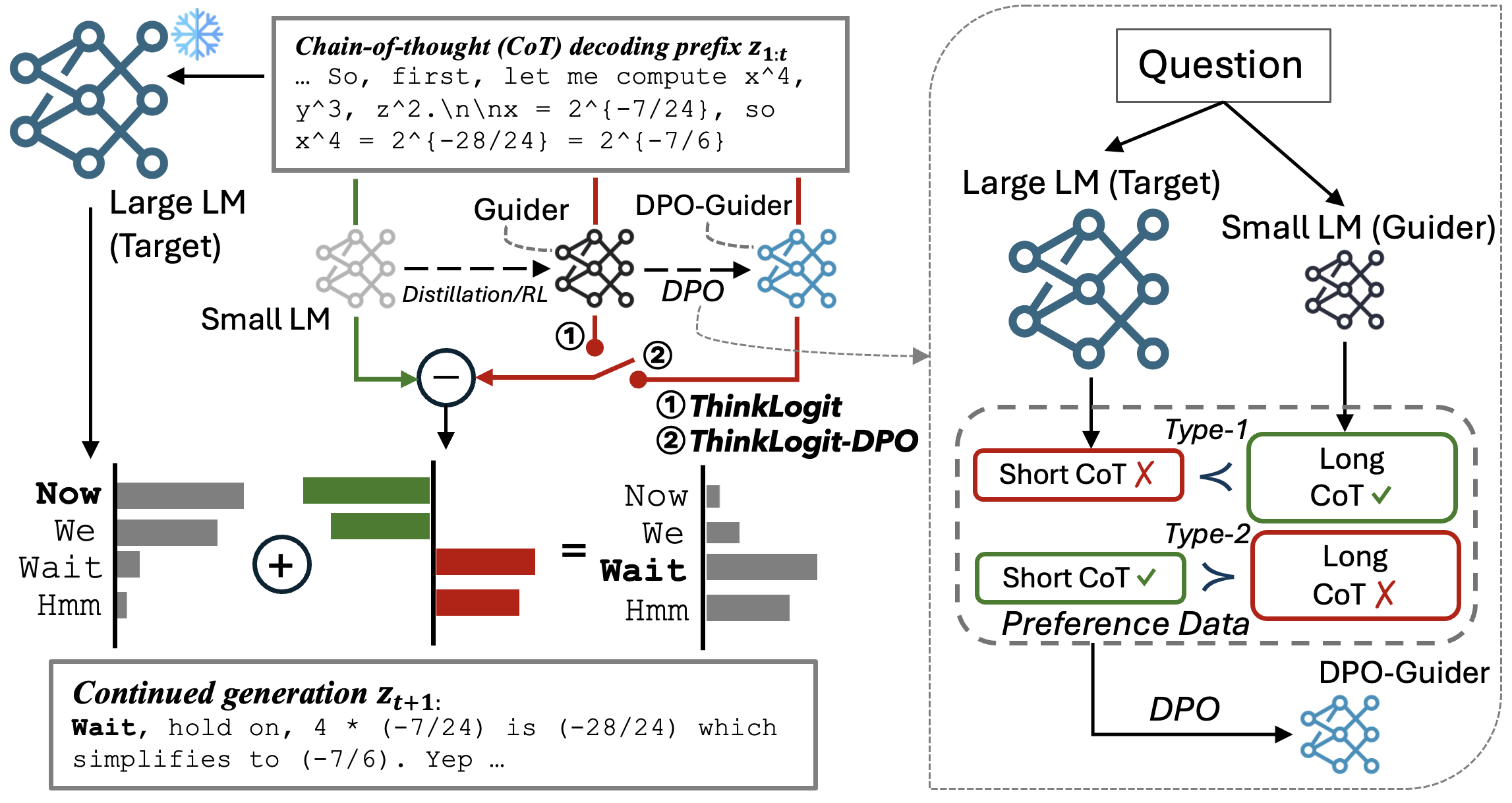}  
  \caption{Overview of our proposed  \mybasemethod and \mymethod approaches to elicit long chain-of-thought reasoning from a large non-reasoning model that is frozen.
  }
  \label{fig:overview}
\end{figure*}

\paragraph{Logit Arithmetic.}
Logit arithmetic blends output distributions from multiple models to steer generation~\citep{liu2021dexperts,DBLP:conf/emnlp/OrmazabalAA23,DBLP:conf/nips/FanL0TQCC24,DBLP:conf/naacl/ShiHLTZY24}, and has been applied to emulating pretraining~\citep{mitchellemulator}, task-specific fine-tuning~\citep{liutuning,DBLP:conf/nips/FanL0TQCC24}, knowledge unlearning~\citep{DBLP:journals/tmlr/HuangZ0M0PC25}, and overriding safety filters~\citep{DBLP:journals/corr/abs-2401-17256}. 
However, prior applications typically target \textit{local} attributes like vocabulary or style. 
In contrast, our setting demands modeling \textit{long-range dependencies} to sustain complex cognitive behaviors like backtracking and verification over extended sequences~\citep{DBLP:journals/corr/abs-2503-01307}. 
While \textit{concurrent} work~\citep{DBLP:journals/corr/abs-2506-15710} also applies logit arithmetic to reasoning, they rely on direct fusion within the same model family. 
We advance this by introducing DPO alignment for targeted error correction and demonstrating robustness in \textit{cross-family} transfer.

\section{Methodology}
\label{sec:method}
In this section, we introduce \textbf{\mybasemethod} (Section~\ref{sec:thinklogit}) and \textbf{\mymethod} (Section~\ref{sec:thinklogit-dpo}) to elicit long CoT in frozen  non-reasoning models via logit arithmetic and DPO.

\subsection{\mybasemethod}\label{sec:thinklogit}
Let  
\(
\mathbf{z}_{1:t}=z_1,\dots,z_t
\)  
be the partially decoded sequence of reasoning tokens at step \(t\).  
For any language model \(f\), denote its pre-softmax logits at the next step by
\(
\mathbf{\ell}_{t+1}^{(f)}\! = f(\mathbf{z}_{1:t})
\in\mathbb{R}^{|\mathcal{V}|},
\)
where \(\mathcal{V}\) is the vocabulary.
\noindent We assume three models during inference:
\begin{itemize}[itemsep=2pt, topsep=2pt, parsep=0pt, partopsep=0pt]
    \item \textbf{large (target)} $L$, a pre-trained LLM lacking long CoT capability;
    \item \textbf{small base} $S_0$, a pre-trained model without long reasoning fine-tuning;
    \item \textbf{small reasoning (guider)} $S$, obtained via long CoT post-training to $S_0$.
\end{itemize}

As shown in Figure~\ref{fig:overview}, at decoding step \(t{+}1\), the fused logits are computed as $
\mathbf{\tilde\ell}_{t+1}
  = \mathbf{\ell}_{t+1}^{(L)}
  \;+\;
    \alpha \cdot \bigl(
        \mathbf{\ell}_{t+1}^{(S)}
      - \mathbf{\ell}_{t+1}^{(S_0)}
    \bigr),
$
where \(\alpha \ge 0\) controls the guidance strength.  
The delta term
\(
\mathbf{\ell}^{(S)}-\mathbf{\ell}^{(S_0)}
\)
encodes the probability shift that turns a short-CoT model into a long-CoT one.
In the illustrated example, while the target model favors shallow continuations like ``Now'' or ``We,'' \mybasemethod\ shifts the top prediction to ``Wait'' by fusing the reasoning-oriented delta logits. This intervention triggers a reflective self-correction step (``\textit{Wait, hold on...}'').
In Appendix~\ref{app:theory}, we provide a theoretical analysis justifying that our logit arithmetic formulation behaves approximately linearly across tasks.

\subsection{\mymethod}\label{sec:thinklogit-dpo}
The effectiveness of \mybasemethod can be constrained by distribution mismatches between the guider and target. To address this, we further train the small model as a stronger guider that corrects target reasoning errors while retaining the target strengths, using a mixture of two types of preference pairs sampled from both the target $L$ and the guider $S$ outputs (see Figure~\ref{fig:overview}):

\begin{description}[leftmargin=1.4em]
  \item[Type-1:] $(x,\,y^{L\checkmark},\,y^{S\times})$  
        — The \textit{large} model’s correct (short) CoT is preferred over the \textit{small} model’s incorrect (long) one.  
        This encourages the guider to preserve the correctness of the target model and avoid introducing new errors.
  \item[Type-2:] $(x,\,y^{S\checkmark},\,y^{L\times})$  
        — The \textit{small} model’s correct (long) CoT is preferred over the \textit{large} model’s incorrect (short) one, teaching the guider to be more confident at fixing the large model’s reasoning errors.
\end{description}

We gather these pairs from training queries \(x\) by independently sampling CoTs from \(L\) and \(S\) and labeling correctness based on the final answer.
Let \(\theta\) denote the parameters of the preference-optimized guider, initialized from \(S\).  
We train \(\theta\) with a DPO objective function that mixes the two pair types:

\begin{table*}
  \centering
  \small
  \setlength{\tabcolsep}{3pt}  
    \begin{tabular}{lc|llllll}
    \toprule
    Model & \makecell[c]{\# Trainable\\Params} & \makecell[l]{AIME\\2024} & \makecell[l]{AIME\\2025} & \makecell[l]{AMC\\23} & \makecell[l]{MATH\\Level 5} & \makecell[l]{GPQA\\Diamond} & \makecell[l]{LiveCode\\Bench v5} \\
    \midrule
    \textbf{\texttt{(Guider)}} R1-Distill-Qwen-1.5B                          & -     & 16.2 & 18.8 & 51.2 & 47.5 & 28.9 & 36.2 \\
    \textbf{\texttt{(Target)}} Qwen2.5-32B                                   & -     & 14.6 &  8.3 & 57.2 & 44.7 & 36.9 & 48.3  \\\midrule
    \textbf{\textit{No Fine-tuning of the \texttt{Target}}}\\
    \textbf{\texttt{Target}} \colorbox{modelblue}{\,+ \mybasemethod\,}     & 0      & 22.5\mydelta{6.3} & 19.2\mydelta{0.3} & 62.2\mydelta{5.0} & 55.3\mydelta{7.8} & 41.8\mydelta{4.9} & 54.1\mydelta{5.8} \\
    \textbf{\texttt{Target}} \colorbox{modelblue}{\,+ \mymethod\,}   & 78M      & 22.1\mydelta{5.9} & 21.7\mydelta{2.9} & 63.7\mydelta{6.5} & 58.5\mydelta{11.0}  & 42.4\mydelta{5.5} & 52.4\mydelta{4.1}  \\\midrule
        \textbf{\textit{Full Fine-tuning of the \texttt{Target}}}\\
        s1.1-32B            & 32B  & 32.9 & 25.4 & 70.0 & 72.2 & 51.9  & 58.0  \\
    R1-Distill-Qwen-32B             & 32B  & 45.8 & 35.0 & 76.9 & 72.7 & 55.6 & 75.4  \\
    \bottomrule
  \end{tabular}

  \caption{
Comparison of avg@8 performance across six reasoning benchmarks covering math, science, and coding domains. The \colorbox{deltayellow}{\,yellow cells\,} highlight the improvement of our methods over the stronger baseline model (\textbf{\texttt{Target}} or \textbf{\texttt{Guider}}). 
We show that \mybasemethod and \mymethod provide substantial gains over the baselines and partially recovers the benefits of full-model fine-tuning without any training of the large target model.
}
  \label{tab:reasoning-transfer-results}
\end{table*}

\begin{equation}\label{eq:dpo}
\begin{aligned}
\mathcal{L}_{\text{DPO}}(\theta)=
\lambda\,
  \mathbb{E}_{(x,y^{L\checkmark},y^{S\times})\sim\mathcal{D}_1}
  \,\ell_\theta\!\bigl(x;\,y^{L\checkmark},y^{S\times}\bigr)\\
+(1-\lambda)\,
  \mathbb{E}_{(x,y^{S\checkmark},y^{L\times})\sim\mathcal{D}_2}
  \,\ell_\theta\!\bigl(x;\,y^{S\checkmark},y^{L\times}\bigr),
\end{aligned}
\end{equation}
where 
$
\ell_\theta\!\bigl(x;\,y^{+},y^{-}\bigr)
=\log\sigma\!\bigl(r_\theta(x,y^{+})-r_\theta(x,y^{-})\bigr)
$,
\(\sigma\) is the sigmoid function,  
$
  r_\theta(x,y)=\beta\!\left[
        \log\pi_\theta (y\mid x)
        -\log\pi_{\text{ref}}(y\mid x)
      \right]
$ is the implicit reward of output \(y\), and  
\(\lambda\in[0,1]\) balances the two datasets \(\mathcal{D}_1\) (Type-1) and \(\mathcal{D}_2\) (Type-2).  
We use \(\lambda{=}\frac{|\mathcal{D}_1|}{|\mathcal{D}_1|+|\mathcal{D}_2|}\) by default, 
directly concatenating two datasets as DPO training data without further rebalancing.
After fine-tuning, we replace \(S\) in \mybasemethod with the optimized guider to obtain \mymethod.

\section{Experimental Setup}

\paragraph{Benchmarks.}
We evaluate models on six widely used  reasoning benchmarks for LRMs. Four of them are competition math problems sources from \textsc{AIME2024}, \textsc{AIME2025}, \textsc{AMC23}, and Level 5 hard problems from MATH-500~\citep{DBLP:conf/iclr/LightmanKBEBLLS24}. 
For scientific reasoning, we evaluate on GPQA Diamond~\citep{DBLP:journals/corr/abs-2311-12022}, consisting of 198 PhD-level science questions in Biology, Chemistry, and Physics.
For code reasoning, we use a 200-problem subset from the v5 release of LiveCodeBench~\citep{DBLP:conf/iclr/JainHGLYZWSSS25}. 
For each dataset, we independently sample 8 completions with a decoding temperature of 0.6 and maximum output length of 8192, and then compute their average accuracy (for math and science reasoning) or pass rate (for code reasoning) as \textbf{Avg@8} for our primary metric.

\paragraph{Models.} Our primary setup pairs a \textbf{Qwen2.5-32B}~\citep{DBLP:journals/corr/abs-2412-15115} target with an \textbf{R1-Distill-Qwen-1.5B}~\citep{DBLP:journals/corr/abs-2501-12948} guider. The guider is initialized from Qwen2.5-Math-1.5B~\citep{DBLP:journals/corr/abs-2409-12122} and fine-tuned on distilled long CoTs from DeepSeek-R1~\citep{DBLP:journals/corr/abs-2501-12948}. The guider and target share the same Qwen tokenizer, which enables direct application of logit arithmetic. To demonstrate versatility, we also evaluate: (1) \textit{Cross-family targets} (Section~\ref{sec:cross_family}): We apply the Qwen-based guider to large non-reasoning models from three distinct model families: \textbf{Llama-3.3-70B-Instruct}~\citep{DBLP:journals/corr/abs-2407-21783} built by Meta, \textbf{OLMo-2-0325-32B-Instruct}~\citep{DBLP:journals/corr/abs-2501-00656} built by Ai2, and \textbf{EXAONE-3.5-32B-Instruct} built by LG AI Research~\citep{DBLP:journals/corr/abs-2412-04862}. (2) \textit{RL-trained guiders} (Section~\ref{sec:emulating_rl}): We test guiders produced via RL rather than supervised distillation: \textbf{One-Shot-RLVR-1.5B}~\cite{wang2025reinforcement} and \textbf{DeepScaleR-1.5B-Preview}~\cite{luo2025deepscaler}. We set the guidance strength to $\alpha=1$ by default.

\paragraph{Preference Data Construction.}
We construct our preference dataset using the level 4–5 subset of the MATH training set~\citep{DBLP:conf/nips/HendrycksBKABTS21}. For each question, we sample completions from both the guider and target models to construct conflicting pairs where one model is correct and the other is incorrect. From the resulting pool of over 55k candidate pairs (spanning both Type-1 and Type-2 categories), we randomly select a subset of 10k pairs for DPO training. To save training compute, we apply LoRA~\citep{DBLP:conf/iclr/HuSWALWWC22} with a rank size of 64 for parameter-efficient fine-tuning of the guider model. Detailed statistics regarding completion counts and pair distributions are provided in Appendix~\ref{subsec:training-details}.

\section{Experiment Results}
\label{sec:exp}
\subsection{Main Results}
\label{sec:results}

Table~\ref{tab:reasoning-transfer-results} presents the avg@8 accuracies for all systems. We highlight two key observations.
\textit{
First, \mybasemethod consistently enhances reasoning performance across math, science, and coding benchmarks, outperforming both the target and guider models.}
Combining the logits of the 32B target with those of the 1.5B guider (\mybasemethod) raises the average performance by 21.5\% relative to the frozen target and by 28.3\% relative to the guider.
While \mymethod performs better than \mybasemethod by 3.7\% relative in math, the improvement does not transfer to code reasoning. We attribute this to the fact that the DPO-trained guider was optimized exclusively on mathematical preference pairs. These results highlight that preference alignment is domain-dependent, suggesting that multi-domain training data is necessary to generalize these benefits to other reasoning tasks.

\textit{Second, our approach recovers a significant portion of the performance benefits of full-parameter fine-tuning without any training.} 
Remarkably, \mybasemethod bridges 45\% of the performance gap to the data-efficient s1.1-32B~\citep{DBLP:journals/corr/abs-2501-19393} baseline (trained on 1K examples) and 30\% of the gap to the stronger R1-Distill-Qwen-32B~\citep{DBLP:journals/corr/abs-2501-12948} upper bound (trained on 800K examples).
It achieves this by leveraging an off-the-shelf 1.5B guider, requiring zero gradient updates or data collection. This demonstrates that a substantial fraction of the reasoning capabilities typically acquired through expensive full-model training can instead be elicited at inference time by simply reusing existing open-weights reasoning models.

\begin{figure}
  \centering
  \includegraphics[width=\linewidth]{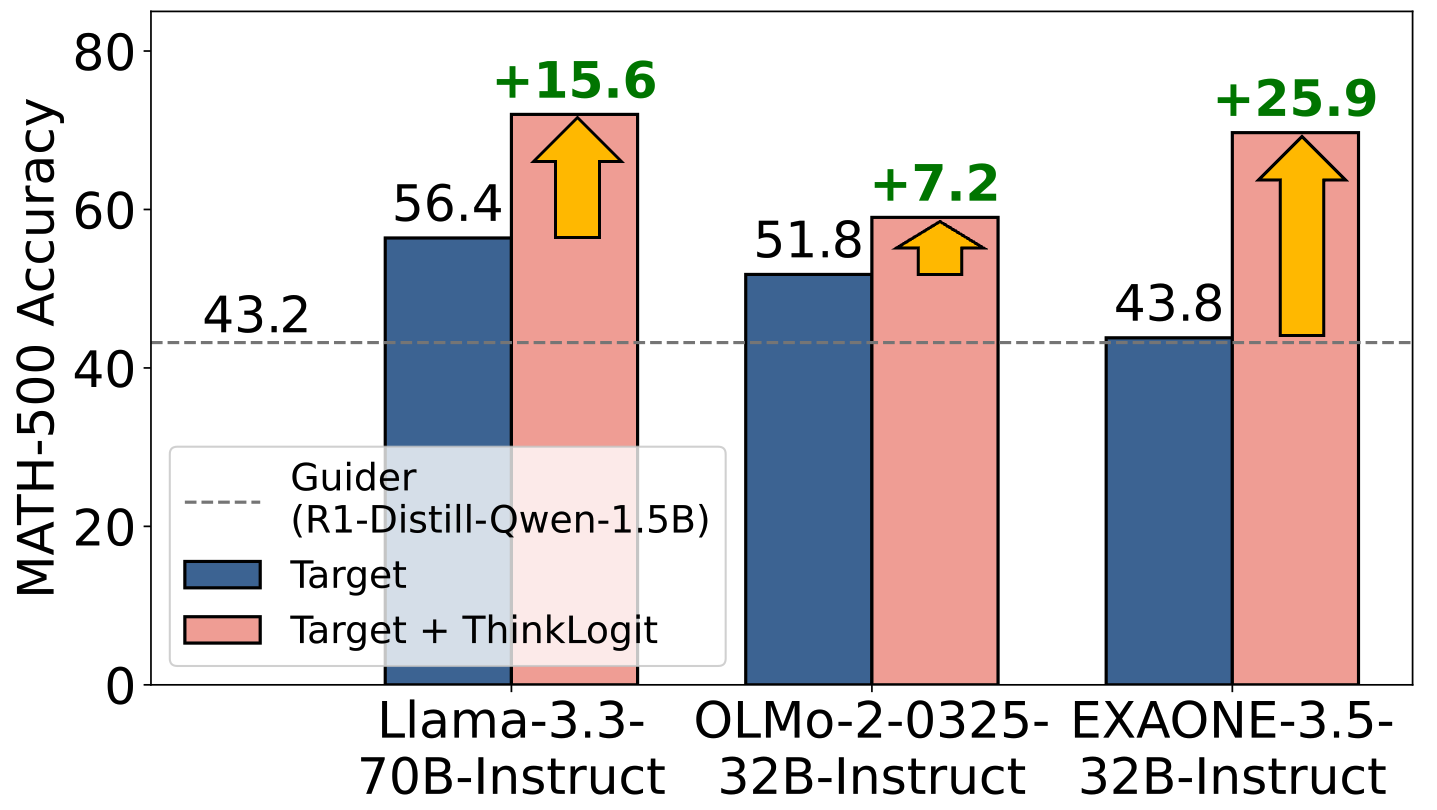}  
\caption{Cross-family long reasoning elicitation on MATH-500~\citep{DBLP:conf/iclr/LightmanKBEBLLS24} , applying a Qwen-based guider to Llama, OLMo, and EXAONE targets with greedy decoding. \mybasemethod generalizes effectively across heterogeneous model families, delivering consistent accuracy gains (up to +25.9\% absolute) despite architectural and tokenizer differences.}
  \label{fig:cross_family}
\end{figure}
\subsection{Generalization Across Model Families} \label{sec:cross_family} 
In our main results, both the guider and target models come from the same Qwen model family. We now show that the small reasoning guider can also be applied to target models from  different families. Although logit arithmetic typically assumes a shared tokenizer, we overcome this constraint by aligning vocabularies with a minimum edit distance mapping~\cite{DBLP:conf/iclr/WanH0QB024}. In practice, this is a one-time, offline step that introduces no inference overhead.

Using this strategy, we employ the R1-Distill-\textbf{Qwen}-1.5B to guide three target models: \textbf{Llama}-3.3-70B-Instruct, \textbf{OLMo}-2-0325-32B-Instruct, and \textbf{EXAONE}-3.5-32B-Instruct. As shown in Figure~\ref{fig:cross_family}, \mybasemethod consistently improves reasoning performance across all three heterogeneous targets on the MATH-500 benchmark, with the most notable gains observed on EXAONE-3.5-32B-Instruct, which achieves a 25.9\% absolute accuracy increase. 
These results highlight the broad applicability of our approach: \textit{a single specialized guider can be reused for larger models within or across families}, offering a scalable way to unlock long reasoning without retraining each target.

\begin{table}
\centering
\setlength{\tabcolsep}{3pt}
\small
\begin{tabular}{lll}
\toprule
Model  & \makecell[l]{AIME\\2024} & \makecell[l]{AIME\\2025} \\
\midrule
\multicolumn{3}{l}{\textbf{\textit{Emulating RL on a Base \texttt{Target}}}}\\
\textbf{\texttt{(Guider)}} One-Shot-RLVR-1.5B  & 13.3  & 7.1 \\
\textbf{\texttt{(Target)}} Qwen2.5-32B   & 14.6 &  8.3 \\
\textbf{\texttt{Target}} \colorbox{modelblue}{\,+ \mybasemethod\,}   & 17.5\mydelta{2.9} & 11.2\mydelta{2.9} \\\midrule
\multicolumn{3}{l}{\textbf{\textit{Emulating RL on a Supervised Fine-Tuned \texttt{Target}}}}\\
\textbf{\texttt{(Guider)}} DeepScaleR-1.5B  & 30.0  & 23.8 \\
\textbf{\texttt{(Target)}} R1-Distill-Qwen-32B   & 45.8  & 35.0 \\
\textbf{\texttt{Target}} \colorbox{modelblue}{\,+ \mybasemethod\,}   & 47.5\mydelta{1.7}  & 37.9\mydelta{2.9} \\
\bottomrule
\end{tabular}
\caption{
Avg@8 performance of \mybasemethod emulating reinforcement learning (RL) on large target models.
The two small guiders are trained via RL.
Both emulated RL pipelines deliver consistent performance gains while avoiding the prohibitive cost of applying RL training on large target models.
}
\label{tab:rl_performance}
\end{table}

\subsection{Emulation of Reinforcement Learning}
\label{sec:emulating_rl}

RL with verifiable rewards is a powerful paradigm for enhancing the reasoning capabilities of language models~\cite{DBLP:journals/corr/abs-2411-15124,DBLP:journals/corr/abs-2402-03300}, but it is often prohibitively expensive. For example, training even a small 1.5B model with RL for long reasoning can require thousands of  GPU hours~\cite{deepscaler2025}, an expense that becomes prohibitive for large-scale models. In Table~\ref{tab:rl_performance}, we investigate whether \mybasemethod could emulate the effects of RL on a large target model without training it.

First, we simulate a \textbf{Zero-RL}~\citep[applying RL directly on a base model;][]{DBLP:journals/corr/abs-2501-12948,DBLP:journals/corr/abs-2503-18892} pipeline to enhance a large non-reasoning model, Qwen2.5-32B. We apply reasoning from the One-Shot-RLVR-1.5B~\cite{wang2025reinforcement} guider, which is RL-trained on only one example from a Qwen2.5-Math-1.5B base.
Second, we simulate an \textbf{SFT-then-RL} pipeline to further improve an already supervised fine-tuned reasoning model like R1-Distill-Qwen-32B. We apply guidance from the 
DeepScaleR-1.5B-Preview~\cite{deepscaler2025} guider, which is trained via distributed RL from an R1-Distill-Qwen-1.5B base.

Table~\ref{tab:rl_performance} shows that both emulated pipelines deliver consistent performance gains over the target large model and RL-trained small guiders, while \textit{avoiding the prohibitive cost of applying RL training directly to the large target model}. 
This confirms that our method is an \textit{orthogonal} technique that can directly benefit from advances in small-model post-training (e.g., distillation or RL), \textit{offering a flexible and efficient mechanism to transfer the benefits of powerful but expensive training paradigms to larger models}.

\begin{figure}
  \centering
  \includegraphics[width=\linewidth]{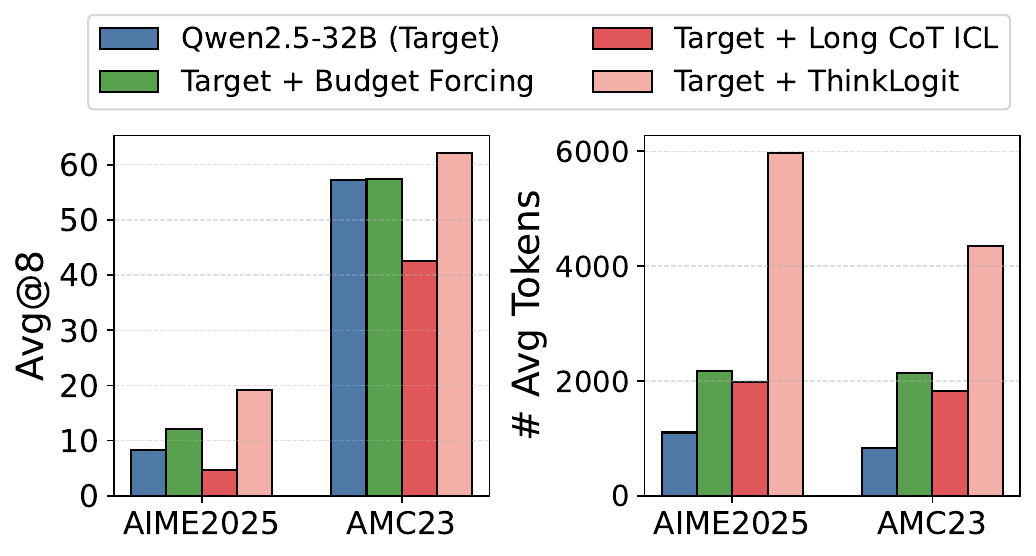} 
  \caption{Comparison of \mybasemethod against two training-free long CoT elicitation baselines: budget forcing and one-shot long CoT in-context learning (ICL). While these approaches increase verbosity, their accuracies are generally lower and can even degrade, whereas \mybasemethod consistently produces longer reasoning that delivers the best performance.}
  \label{fig:elicit}
\end{figure}

\subsection{Comparison with Training-Free Baselines}\label{sec:elict}

A natural question is whether existing training-free techniques are sufficient to elicit long CoTs, or if specialized methods are required. Figure~\ref{fig:elicit} contrasts our approach against two such baselines.
First, the budget-forcing\footnote{For budget forcing, we report results using Qwen2.5-32B-Instruct, since applying it to the base Qwen2.5-32B led to low-quality outputs and degraded performance. This choice is consistent with the setup in~\citet{DBLP:journals/corr/abs-2501-19393}.} heuristic~\citep{DBLP:journals/corr/abs-2501-19393} replaces the end-of-sentence token with a placeholder string like ``Wait'' to artificially increase output length. Crucially, we find that further scaling the number of tokens in this baseline yields no additional accuracy gains; instead, the model frequently degenerates into repetitive loops—restating the same conclusion without producing new information—rather than generating valid cognitive steps. In contrast, our method elicits genuine backtracking and verification, confirming that \textit{mere verbosity does not translate into deeper reasoning}.

Second, inserting a single long CoT example sourced from the s1.1‑1K dataset~\citep{DBLP:journals/corr/abs-2501-19393} into the prompt for in-context learning~\citep[ICL;][]{DBLP:conf/nips/BrownMRSKDNSSAA20,DBLP:conf/emnlp/MinLHALHZ22,DBLP:conf/emnlp/Dong0DZMLXX0C0S24} also degrades performance despite longer outputs from the target model. In contrast, \mybasemethod uses logit‑level guidance from a small reasoning model to steer the decoding towards genuine long CoT reasoning, which translates into a clear uplift in the answer accuracy. This shows that our improvements stem from \textit{the quality of the guidance} being applied, rather than \textit{the quantity of tokens} generated.

\begin{figure}
  \centering
  \includegraphics[width=\linewidth]{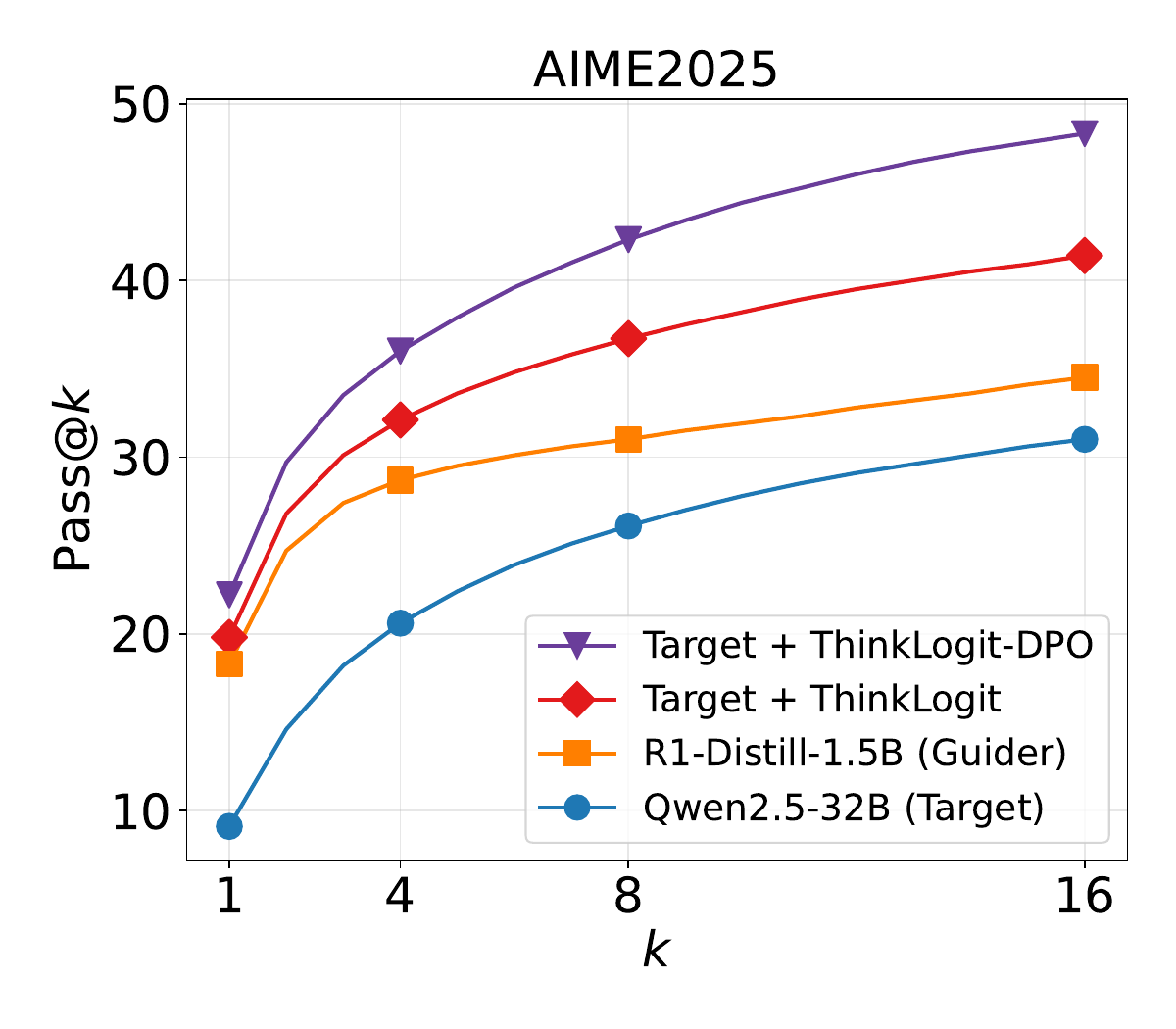}  
\caption{Inference‑time scaling on AIME2025.  \textit{Pass@\,$k$} for $k\!=\!1\text{--}16$ comparing the target, guider, their direct logit fusion (\mybasemethod), and the DPO‑aligned fusion (\mymethod). Our methods demonstrate superior sample efficiency, reaching stronger performance with fewer generations and maintaining larger gains as the sample budget increases.}
  \label{fig:pass_at_k}
\end{figure}

\subsection{Inference-Time Scaling Properties}\label{app:test_time_scaling}

We further evaluate the \textit{scalability} of \mybasemethod concerning both repeated sampling and  guider capacity. Figure~\ref{fig:pass_at_k} shows that \mybasemethod scales effectively with the number of generations per question.
Test-time scaling via repeated sampling is a well-established strategy for enhancing reasoning performance~\cite{DBLP:journals/corr/abs-2407-21787,DBLP:journals/corr/abs-2408-03314}, as drawing more samples naturally increases the probability of finding a correct solution. We quantify this effect using the pass@$k$ metric~\citep{DBLP:journals/corr/abs-2107-03374}, which measures the likelihood that at least one of $k$ sampled outputs is correct. Figure~\ref{fig:pass_at_k} plots pass@$k$ for $k=1\text{--}16$ on AIME2025, the benchmark where the 32B target is weakest and scaling effects are most pronounced.

Notably, both \mybasemethod{} and \mymethod{} surpass the target's pass@16 accuracy using only four samples, demonstrating a fourfold increase in \textit{sample efficiency}. This performance gap widens as $k$ scales: at $k=16$, the DPO-aligned guider outperforms the target by roughly 17 points. Ultimately, these results confirm that inference-time guidance reliably improves reasoning, with gains that compound alongside larger sampling budgets.

Although our primary configuration uses a 1.5B guider for efficiency, we demonstrate in Appendix~\ref{app:guider_scaling} that performance consistently improves as guider size increases.

\section{Additional Analyses}

\subsection{Ablation Study of \mymethod}\label{sec:dpo-ablation}

To further investigate the design choices in \mymethod{}, we ablate both our mixed‑pair data construction and preference‑based learning objective (DPO) against single‑source or supervised fine-tuning alternatives. Results in Table~\ref{tab:decoding_ablation_additional} answer the following research questions.

\begin{table}
  \centering
  \small
  \setlength{\tabcolsep}{3pt}
  \begin{tabular}{lcc}
    \toprule
    Model & \makecell[c]{Training Data\\for Guider} & Avg@8\\
    \midrule
    \rowcolor{lightgray}
    \multicolumn{3}{c}{\textsc{\textbf{ThinkLogit-DPO}}}\\\midrule
        Ours 
      & \makecell[c]{$(x,\,y^{L\checkmark},\,y^{S\times})$,\\$(x,\,y^{S\checkmark},\,y^{L\times})$}  & \textbf{63.7} \\
    \quad w/o dual sources & $(x,\,y^{S\checkmark},\,y^{S\times})$                      
      & 58.8 \\
    \quad w/o Type-1 pairs & $(x,\,y^{S\checkmark},\,y^{L\times})$                    
      & 57.2  \\
    \quad w/o Type-2 pairs & $(x,\,y^{L\checkmark},\,y^{S\times})$                   
      & 51.9 \\\midrule
    \rowcolor{lightgray}
    \multicolumn{3}{c}{\textsc{\textbf{ThinkLogit-SFT}}}\\\midrule
    learning from target & $(x, y^{L\checkmark})$
      & 44.7 \\
    self-learning &  $(x, y^{S\checkmark})$
      & 55.6 \\
    learning from teacher & $(x, y^{R1\checkmark})$
      & 60.9 \\
    \bottomrule
  \end{tabular}
  \caption{Avg@8 on AMC23 under ablations of guider's training data and objectives in \mymethod. 
  $x,\,y^{L\checkmark},$ and $y^{S\times}$ denote the question, the correct ($\checkmark$) response for the large target model ($L$), and the incorrect ($\times$) response from the small guider model ($S$), respectively.
  Our dual-source, mixed-pair DPO performs the best, demonstrating the necessity of complementary preference signals and preference-based alignment.}

  \label{tab:decoding_ablation_additional}
\end{table}

\paragraph{Are preference pairs sourced from both the target and the guider necessary to maximize performance?}
We construct the same amount of 10K preference pairs using only the guider’s correct vs. incorrect outputs, i.e., $(x,\,y^{S\checkmark},\,y^{S\times})$. DPO on this data underperforms markedly on AMC23 (58.8 vs.\ 63.7 by our \mymethod), confirming that mixing pairs which highlight \emph{both} the target’s and guider’s strengths is crucial for maximal gains.

\paragraph{Is training on both types of pairs necessary for the effectiveness of \mymethod?}
We next ablate by training on only one type of preference pairs at a time: using only Type‑2 pairs $(x,\,y^{S\checkmark},\,y^{L\times})$ (i.e., $\lambda=0$ in Equation~\ref{eq:dpo}) yields an avg@8 of 57.2, while using only Type‑1 pairs $(x,\,y^{L\checkmark},\,y^{S\times})$ (i.e., $\lambda=1$ in Equation~\ref{eq:dpo}) drops further to 51.9.
Both are substantially below the 63.7 achieved by the full mixture, indicating that \textit{both} Type‑2 pairs (which teach the guider to correct target errors) and Type‑1 pairs (which enforce preservation of the correct  reasoning of the target) are necessary for better alignment.

\paragraph{Can supervised fine-tuning replace preference-based alignment of the guider?}  
We evaluate SFT against DPO by training the guider on three equally sized sets of high-quality completions:  
\textbf{Option 1:} the target model’s correct outputs \(y^{L\checkmark}\);  
\textbf{Option 2:} the guider’s own correct outputs \(y^{S\checkmark}\) (also known as rejection-sampling fine-tuning~\citep{DBLP:journals/corr/abs-2308-01825});  
\textbf{Option 3:} R1-distilled completions \(y^{R1\checkmark}\).  
Although SFT on Options 1 and 2 makes the guider a stronger \textit{standalone} reasoner, none of these variants match the performance of the DPO-aligned guider. This gap highlights that \textit{optimizing with pairwise preference comparisons yields a better guider than optimizing solely for correctness}.  
While SFT on Option 3 adapts the guider toward the target’s short CoT reasoning style and thus reduces the distributional gap, it also tends to overwrite the guider’s native strengths of long reasoning. In contrast, DPO \textit{preserves the guider’s intrinsic reasoning capabilities} while \textit{selectively aligning it to the target’s preferences} through pairwise comparisons.

We also ablate the guidance strength $\alpha$ in Appendix~\ref{app:hyperparam}, where we observe that the default value $\alpha=1.0$ performs best. We applied the exact same hyperparameter configuration across all main experiments, encompassing six benchmarks and six guider-target model pairs. This required zero per-task or per-model tuning, confirming the strong out-of-the-box robustness of our method.

\subsection{Reasoning Behavior Analysis}
\label{sec:behavior}

To verify that \mybasemethod and \mymethod elicit genuine long reasoning capabilities rather than simply increasing output verbosity, we analyze the frequency of tokens associated with three key reasoning behaviors: \textbf{Branching out} (exploring alternative approaches), \textbf{Backtracking} (correcting errors or revising steps), and \textbf{Self-verification} (systematic checking of intermediate results).

To quantify these behaviors, we curate a lexicon of representative keywords based on prior studies~\citep{DBLP:journals/corr/abs-2503-01307,DBLP:journals/corr/abs-2506-15710}. We process the model reasoning traces with the spaCy~\citep{honnibal2020spacy} toolkit using tokenization and lemmatization to account for morphological variations, then calculate the frequency of keywords associated with each reasoning behavior. The full list of keywords used for this analysis is provided in Table~\ref{tab:keywords} of Appendix~\ref{app:behavior}.

\begin{figure}
  \centering
  \includegraphics[width=\linewidth]{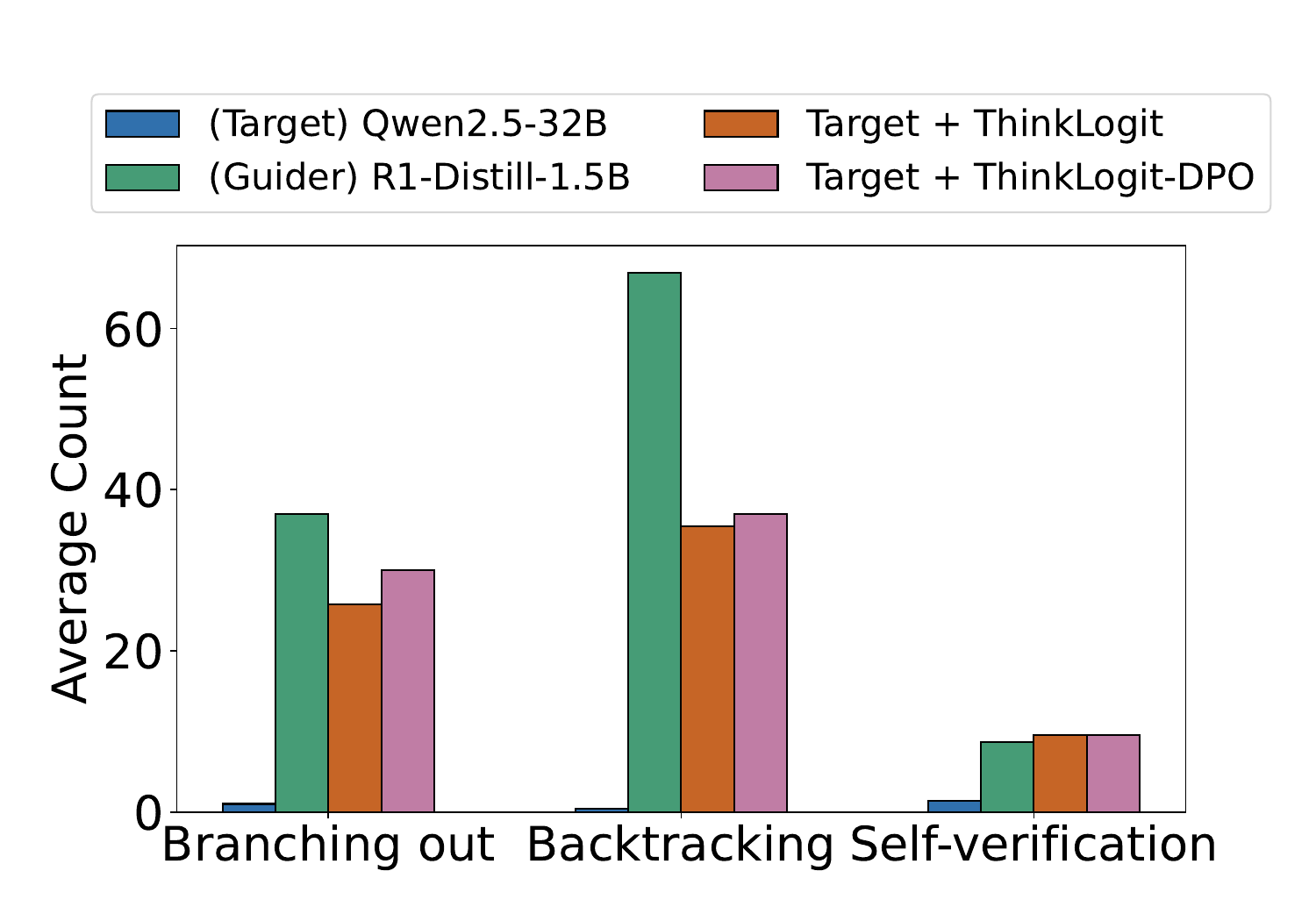}  
\caption{Average frequency of key reasoning behaviors (Branching out, Backtracking, and Self-verification) across all six benchmarks. \mybasemethod and \mymethod successfully elicit complex cognitive patterns in the target non-reasoning model, effectively transferring the long reasoning capabilities of the small guider.}
  \label{fig:reasoning_behavior}
\end{figure}

We present the average frequency of these behaviors across all six datasets in Figure~\ref{fig:reasoning_behavior}, which reveals a stark contrast before and after applying \mybasemethod. The target model (Qwen2.5-32B) exhibits negligible occurrences of these reasoning behaviors, which is consistent with its inability to perform long chain-of-thought reasoning. In contrast, the guider model (R1-Distill-1.5B) demonstrates a high frequency of reasoning-related tokens, particularly for Backtracking and Branching out. \mybasemethod and \mymethod successfully transfer these behaviors to the large target model, eliciting substantial amounts of branching, backtracking, and self-verification, effectively bridging the gap between the non-reasoning target and the specialized guider.
We further break down these behaviors by domain in Figure~\ref{fig:reasoning_behaviour_combined} (see Appendix~\ref{app:behavior}). The results show consistent trends across math, science, and coding tasks. 

\begin{figure*}[h]
\footnotesize
\centering
\begin{tabular}{@{}p{0.29\textwidth}p{0.34\textwidth}p{0.31\textwidth}@{}}
\toprule

\multicolumn{3}{p{\linewidth}}{\textbf{Question:} \textit{Let $P(x)$ be a monic polynomial of degree 3.  Suppose that $P(x)$ has remainder $R(x)$ when it is divided by $(x - 1)(x - 4),$ and remainder $2R(x)$ when it is divided by $(x - 2)(x - 3).$  Given that $P(0) = 5,$ find $P(5).$} \newline
\textbf{Answer:} 15.}\\\midrule
\textbf{(\texttt{Guider}) R1-Distill-Qwen-1.5B}

~

\input{model_outputs/guider_output}

&
\textbf{(\texttt{Target}) Qwen2.5-32B}

~

\input{model_outputs/target_output}

~

&
\textbf{\texttt{Target} + \mybasemethod}

~

\input{model_outputs/thinklogit_output}

\\
\bottomrule
\end{tabular}
\vspace{-5pt}
\caption{Example model outputs. We show a question from MATH500, where both the \textbf{\texttt{Guider}} and \textbf{\texttt{Target}} provide \txtred{incorrect answers} but our \mybasemethod generates the \txtgreen{correct answer}. [...] indicates that the text was trimmed to fit this page, but the generated text is actually longer. We highlight text that exhibits long CoT behaviors (e.g., backtracking, branching out, self-verification) in \hl{yellow}. \mybasemethod elicits informative long CoT reasoning behaviors from the non-reasoning target model.}
\label{fig:ex}
\end{figure*}

We provide an example model output after applying \mybasemethod in Figure~\ref{fig:ex}, showing that \mybasemethod elicits rich long CoT behaviors like backtracking and self-verification rather than merely increasing verbosity.
In Appendix~\ref{app:failure_modes}, we further analyze specific failure modes, observing that \mybasemethod can struggle to form higher-level abstractions, hallucinates contradictions that disrupt inductive reasoning, or fail to recover from wrong assumptions.

\begin{table}
\centering
\small
\setlength{\tabcolsep}{2pt}
\begin{tabular}{lccc}
\toprule
& \textbf{\texttt{Target}} & \mybasemethod &  SFT \\
\midrule
\rowcolor{lightgray}
\multicolumn{4}{c}{\textit{\textbf{Training Efficiency}}}\\\midrule
Training Parameters & -- & 0 & 32B \\
\# Training GPUs & -- & 0 & 8$\times$80GB* \\
\midrule
\rowcolor{lightgray}
\multicolumn{4}{c}{\textit{\textbf{Inference Efficiency}}}\\\midrule
Inference Parameters & 32B & 35B & 32B \\
\# Inference GPUs & 2$\times$46GB & 2$\times$46GB & 2$\times$46GB \\
\makecell[l]{Inference Throughput \\(tokens/second)} & 10.1 & 7.6 & 9.1 \\
\bottomrule
\end{tabular}
\caption{Comparison of training and inference efficiency, using Qwen2.5-32B as the Target model. We select R1-Distill-Qwen-32B as a representative model for full-parameter supervised fine-tuning (SFT). Training GPUs required for SFT (*) is estimated based on~\citet{llamafactory}. \mybasemethod trades increased inference latency for \textit{zero} training cost, making long reasoning possible when training the large target is infeasible.}
\label{tab:efficiency}
\end{table}

\subsection{Efficiency Analysis}
\label{sec:efficiency_analysis}

\paragraph{Training Efficiency.}
As illustrated in Table~\ref{tab:efficiency}, \mybasemethod entirely eliminates the training overhead associated with adapting large non-reasoning models for long reasoning. While full parameter supervised fine-tuning (SFT) of the 32B target demands significant computational resources (estimated at eight 80GB GPUs), \mybasemethod requires \textit{zero} training for the target model itself. This makes \mybasemethod distinctively advantageous in resource-constrained settings where training a 32B model is computationally infeasible.

\paragraph{Inference Efficiency.}
Table~\ref{tab:efficiency} further compares the inference metrics of \mybasemethod against the frozen target and the fully fine-tuned baseline. Deploying \mybasemethod requires hosting the large target alongside two smaller models: the base model $S_0$ and the reasoning guider $S$. In our primary setup (a 32B target guided by a 1.5B model), this increases the total parameter count by approximately $1.1\times$ (from 32B to 35B) relative to the target alone. Crucially, this combined footprint still fits within the same hardware configuration ($2\times$ 46GB GPUs), avoiding the need for extra inference hardware. 

Regarding inference throughput (tokens generated per second), profiling on NVIDIA L40S GPUs shows a moderate reduction of roughly 25\% (10.1 vs. 7.6 tokens/second) compared to the target in isolation. Notably, \mymethod introduces no additional overhead beyond \mybasemethod, as it simply replaces the guider with a preference-optimized model of identical size. The observed throughput reduction primarily stems from our prototype implementation based on the Huggingface \texttt{generate()} function, which queries the three models sequentially at each decoding step. In a production environment, logits from all three models could be computed \textit{concurrently} across distributed GPUs, which would largely mitigate this latency and make throughput comparable to that of the standalone target model.

\section{Conclusion}
\label{sec:conclusion}

We present \mybasemethod, a training-free framework eliciting long chain-of-thought via logit guidance from a small reasoner, and its optimized variant \mymethod. Delivering up to 24.2\% gains with a 21$\times$ smaller guider, our approach generalizes across model families and emulates RL benefits. These results establish inference-time guidance as a practical alternative to post-training, enabling modular systems where specialized, lightweight models unlock advanced reasoning in large-scale foundation models.

\section*{Limitations}

\paragraph{Inference Overhead.}
While \mybasemethod\ eliminates the prohibitive cost of training large models, it introduces a modest inference overhead by requiring the simultaneous execution of the small guider and base models. In our current sequential implementation, this results in a throughput reduction of approximately 25\% compared to the standalone target. However, as discussed in Section~\ref{sec:efficiency_analysis}, \textit{this latency is primarily an engineering bottleneck rather than a fundamental one brought by our methodology}. Since the additional model weights fit within the same hardware footprint, production deployments can compute guider logits in parallel to achieve throughput comparable to the target model alone. 

Furthermore, \textit{cross-model versatility multiplies compute savings for training}. Unlike standard fine-tuning, which requires independent training for each target, our approach offers high reusability. We demonstrate that a single 1.5B Qwen guider effectively steers models from diverse families such as Llama, OLMo, and EXAONE (Figure~\ref{fig:cross_family}). This one-to-many capability multiplies the training compute saved across all guided models.

Finally, \textit{new methods that reduce overthinking and yield shorter reasoning traces will directly lower our per-token overhead}.
Recent methods actively reduce overthinking to generate shorter reasoning traces~\citep{DBLP:journals/corr/abs-2502-18600,DBLP:journals/corr/abs-2505-21178,DBLP:journals/corr/abs-2505-22662}. This shift is also visible in practice, with Anthropic substantially lowering Claude's reasoning verbosity between model iterations~\citep{zhou2025deepseek}. Because our overhead scales linearly with token count, these shorter reasoning trajectories naturally shrink the absolute inference cost of our method.

\paragraph{Offline Alignment.}
The guider is aligned with the target via DPO on a fixed set of preference pairs.  This \emph{offline} formulation cannot adapt once deployment uncovers new error patterns or distribution drift.  
Incorporating \emph{online} reinforcement learning~\citep{DBLP:journals/corr/SchulmanWDRK17,DBLP:journals/corr/abs-2402-03300} that updates the guider from streamed on‑policy samples could, in principle, reduce this brittleness.
However, on‑policy RL introduces training efficiency and stability challenges that remain open research problems.

\section*{Acknowledgments}
This work is supported in part by LG AI Research, Cisco Research, National Science Foundation through grant 2046016, Air Force Office of Scientific Research under grant FA9550-22-1-0099, and computational resources and services provided by Advanced Research Computing (ARC), a division of Information and Technology Services (ITS) at the University of Michigan, Ann Arbor.
We thank ARR reviewers for their useful feedback.
We also thank the members of the LAUNCH group at the University of Michigan for their discussions and suggestions.

\bibliography{custom}

@misc{zhou2025deepseek,
  title = {DeepSeek Debrief: >128 Days Later},
  author = {Zhou, Wei and Kourabi, AJ and Patel, Dylan},
  year = {2025},
  month = {jul},
  howpublished = {SemiAnalysis},
  url = {https://newsletter.semianalysis.com/p/deepseek-debrief-128-days-later}
}

@article{DBLP:journals/corr/abs-2505-22662,
  author       = {Feng Luo and
                  Yu{-}Neng Chuang and
                  Guanchu Wang and
                  Hoang Anh Duy Le and
                  Shaochen Zhong and
                  Hongyi Liu and
                  Jiayi Yuan and
                  Yang Sui and
                  Vladimir Braverman and
                  Vipin Chaudhary and
                  Xia Hu},
  title        = {AutoL2S: Auto Long-Short Reasoning for Efficient Large Language Models},
  journal      = {CoRR},
  volume       = {abs/2505.22662},
  year         = {2025},
  url          = {https://doi.org/10.48550/arXiv.2505.22662},
  doi          = {10.48550/ARXIV.2505.22662},
  eprinttype   = {arXiv},
  eprint       = {2505.22662},
  bibsource    = {dblp computer science bibliography, https://dblp.org}
}

@article{DBLP:journals/corr/abs-2505-21178,
  author       = {Mingyang Song and
                  Mao Zheng},
  title        = {Walk Before You Run! Concise {LLM} Reasoning via Reinforcement Learning},
  journal      = {CoRR},
  volume       = {abs/2505.21178},
  year         = {2025},
  url          = {https://doi.org/10.48550/arXiv.2505.21178},
  doi          = {10.48550/ARXIV.2505.21178},
  eprinttype   = {arXiv},
  eprint       = {2505.21178},
  bibsource    = {dblp computer science bibliography, https://dblp.org}
}

@article{DBLP:journals/corr/abs-2502-18600,
  author       = {Silei Xu and
                  Wenhao Xie and
                  Lingxiao Zhao and
                  Pengcheng He},
  title        = {Chain of Draft: Thinking Faster by Writing Less},
  journal      = {CoRR},
  volume       = {abs/2502.18600},
  year         = {2025},
  url          = {https://doi.org/10.48550/arXiv.2502.18600},
  doi          = {10.48550/ARXIV.2502.18600},
  eprinttype   = {arXiv},
  eprint       = {2502.18600},
  bibsource    = {dblp computer science bibliography, https://dblp.org}
}

@article{DBLP:journals/corr/abs-2506-15710,
  author       = {Siru Ouyang and
                  Xinyu Zhu and
                  Zilin Xiao and
                  Minhao Jiang and
                  Yu Meng and
                  Jiawei Han},
  title        = {{RAST:} Reasoning Activation in LLMs via Small-model Transfer},
  journal      = {CoRR},
  volume       = {abs/2506.15710},
  year         = {2025},
  url          = {https://doi.org/10.48550/arXiv.2506.15710},
  doi          = {10.48550/ARXIV.2506.15710},
  eprinttype    = {arXiv},
  eprint       = {2506.15710},
  bibsource    = {dblp computer science bibliography, https://dblp.org}
}

@article{honnibal2020spacy,
  author = {Honnibal, Matthew and Montani, Ines and Van Landeghem, Sofie and Boyd, Adriane},
  doi = {10.5281/zenodo.1212303},
  title = {{spaCy: Industrial-strength Natural Language Processing in Python}},
  year = 2020
}

@inproceedings{DBLP:conf/iclr/JainHGLYZWSSS25,
  author       = {Naman Jain and
                  King Han and
                  Alex Gu and
                  Wen{-}Ding Li and
                  Fanjia Yan and
                  Tianjun Zhang and
                  Sida Wang and
                  Armando Solar{-}Lezama and
                  Koushik Sen and
                  Ion Stoica},
  title        = {LiveCodeBench: Holistic and Contamination Free Evaluation of Large
                  Language Models for Code},
  booktitle    = {The Thirteenth International Conference on Learning Representations,
                  {ICLR} 2025, Singapore, April 24-28, 2025},
  publisher    = {OpenReview.net},
  year         = {2025},
  url          = {https://openreview.net/forum?id=chfJJYC3iL},
  bibsource    = {dblp computer science bibliography, https://dblp.org}
}

@article{DBLP:journals/corr/abs-2502-03860,
  author       = {Bo Pang and
                  Hanze Dong and
                  Jiacheng Xu and
                  Silvio Savarese and
                  Yingbo Zhou and
                  Caiming Xiong},
  title        = {{BOLT:} Bootstrap Long Chain-of-Thought in Language Models without
                  Distillation},
  journal      = {CoRR},
  volume       = {abs/2502.03860},
  year         = {2025},
  url          = {https://doi.org/10.48550/arXiv.2502.03860},
  doi          = {10.48550/ARXIV.2502.03860},
  eprinttype    = {arXiv},
  eprint       = {2502.03860},
  bibsource    = {dblp computer science bibliography, https://dblp.org}
}

@article{DBLP:journals/corr/abs-2408-03314,
  author       = {Charlie Snell and
                  Jaehoon Lee and
                  Kelvin Xu and
                  Aviral Kumar},
  title        = {Scaling {LLM} Test-Time Compute Optimally can be More Effective than
                  Scaling Model Parameters},
  journal      = {CoRR},
  volume       = {abs/2408.03314},
  year         = {2024},
  url          = {https://doi.org/10.48550/arXiv.2408.03314},
  doi          = {10.48550/ARXIV.2408.03314},
  eprinttype    = {arXiv},
  eprint       = {2408.03314},
  bibsource    = {dblp computer science bibliography, https://dblp.org}
}

@article{DBLP:journals/corr/abs-2407-21787,
  author       = {Bradley C. A. Brown and
                  Jordan Juravsky and
                  Ryan Ehrlich and
                  Ronald Clark and
                  Quoc V. Le and
                  Christopher R{\'{e}} and
                  Azalia Mirhoseini},
  title        = {Large Language Monkeys: Scaling Inference Compute with Repeated Sampling},
  journal      = {CoRR},
  volume       = {abs/2407.21787},
  year         = {2024},
  url          = {https://doi.org/10.48550/arXiv.2407.21787},
  doi          = {10.48550/ARXIV.2407.21787},
  eprinttype    = {arXiv},
  eprint       = {2407.21787},
  bibsource    = {dblp computer science bibliography, https://dblp.org}
}

@inproceedings{DBLP:conf/iclr/WanH0QB024,
  author       = {Fanqi Wan and
                  Xinting Huang and
                  Deng Cai and
                  Xiaojun Quan and
                  Wei Bi and
                  Shuming Shi},
  title        = {Knowledge Fusion of Large Language Models},
  booktitle    = {The Twelfth International Conference on Learning Representations,
                  {ICLR} 2024, Vienna, Austria, May 7-11, 2024},
  publisher    = {OpenReview.net},
  year         = {2024},
  url          = {https://openreview.net/forum?id=jiDsk12qcz},
  bibsource    = {dblp computer science bibliography, https://dblp.org}
}

@article{DBLP:journals/corr/abs-2412-04862,
  author       = {LG AI Research and
                  Soyoung An and
                  Kyunghoon Bae and
                  Eunbi Choi and
                  Kibong Choi and
                  Stanley Jungkyu Choi and
                  Seokhee Hong and
                  Junwon Hwang and
                  Hyojin Jeon and
                  Gerrard Jeongwon Jo and
                  Hyunjik Jo and
                  Jiyeon Jung and
                  Yountae Jung and
                  Hyosang Kim and
                  Joonkee Kim and
                  Seonghwan Kim and
                  Soyeon Kim and
                  Sunkyoung Kim and
                  Yireun Kim and
                  Yongil Kim and
                  Youchul Kim and
                  Edward Hwayoung Lee and
                  Haeju Lee and
                  Honglak Lee and
                  Jinsik Lee and
                  Kyungmin Lee and
                  Woohyung Lim and
                  Sangha Park and
                  Sooyoun Park and
                  Yongmin Park and
                  Sihoon Yang and
                  Heuiyeen Yeen and
                  Hyeongu Yun},
  title        = {{EXAONE} 3.5: Series of Large Language Models for Real-world Use Cases},
  journal      = {CoRR},
  volume       = {abs/2412.04862},
  year         = {2024},
  url          = {https://doi.org/10.48550/arXiv.2412.04862},
  doi          = {10.48550/ARXIV.2412.04862},
  eprinttype    = {arXiv},
  eprint       = {2412.04862},
  bibsource    = {dblp computer science bibliography, https://dblp.org}
}

@article{DBLP:journals/corr/abs-2501-00656,
  author       = {Team OLMo and
                  Pete Walsh and
                  Luca Soldaini and
                  Dirk Groeneveld and
                  Kyle Lo and
                  Shane Arora and
                  Akshita Bhagia and
                  Yuling Gu and
                  Shengyi Huang and
                  Matt Jordan and
                  Nathan Lambert and
                  Dustin Schwenk and
                  Oyvind Tafjord and
                  Taira Anderson and
                  David Atkinson and
                  Faeze Brahman and
                  Christopher Clark and
                  Pradeep Dasigi and
                  Nouha Dziri and
                  Michal Guerquin and
                  Hamish Ivison and
                  Pang Wei Koh and
                  Jiacheng Liu and
                  Saumya Malik and
                  William Merrill and
                  Lester James V. Miranda and
                  Jacob Morrison and
                  Tyler Murray and
                  Crystal Nam and
                  Valentina Pyatkin and
                  Aman Rangapur and
                  Michael Schmitz and
                  Sam Skjonsberg and
                  David Wadden and
                  Christopher Wilhelm and
                  Michael Wilson and
                  Luke Zettlemoyer and
                  Ali Farhadi and
                  Noah A. Smith and
                  Hannaneh Hajishirzi},
  title        = {2 OLMo 2 Furious},
  journal      = {CoRR},
  volume       = {abs/2501.00656},
  year         = {2025},
  url          = {https://doi.org/10.48550/arXiv.2501.00656},
  doi          = {10.48550/ARXIV.2501.00656},
  eprinttype    = {arXiv},
  eprint       = {2501.00656},
  bibsource    = {dblp computer science bibliography, https://dblp.org}
}

@article{DBLP:journals/corr/abs-2407-21783,
  author       = {Abhimanyu Dubey and
                  Abhinav Jauhri and
                  Abhinav Pandey and
                  Abhishek Kadian and
                  Ahmad Al{-}Dahle and
                  Aiesha Letman and
                  Akhil Mathur and
                  Alan Schelten and
                  Amy Yang and
                  Angela Fan and
                  Anirudh Goyal and
                  Anthony Hartshorn and
                  Aobo Yang and
                  Archi Mitra and
                  Archie Sravankumar and
                  Artem Korenev and
                  Arthur Hinsvark and
                  Arun Rao and
                  Aston Zhang and
                  Aur{\'{e}}lien Rodriguez and
                  Austen Gregerson and
                  Ava Spataru and
                  Baptiste Rozi{\`{e}}re and
                  Bethany Biron and
                  Binh Tang and
                  Bobbie Chern and
                  Charlotte Caucheteux and
                  Chaya Nayak and
                  Chloe Bi and
                  Chris Marra and
                  Chris McConnell and
                  Christian Keller and
                  Christophe Touret and
                  Chunyang Wu and
                  Corinne Wong and
                  Cristian Canton Ferrer and
                  Cyrus Nikolaidis and
                  Damien Allonsius and
                  Daniel Song and
                  Danielle Pintz and
                  Danny Livshits and
                  David Esiobu and
                  Dhruv Choudhary and
                  Dhruv Mahajan and
                  Diego Garcia{-}Olano and
                  Diego Perino and
                  Dieuwke Hupkes and
                  Egor Lakomkin and
                  Ehab AlBadawy and
                  Elina Lobanova and
                  Emily Dinan and
                  Eric Michael Smith and
                  Filip Radenovic and
                  Frank Zhang and
                  Gabriel Synnaeve and
                  Gabrielle Lee and
                  Georgia Lewis Anderson and
                  Graeme Nail and
                  Gr{\'{e}}goire Mialon and
                  Guan Pang and
                  Guillem Cucurell and
                  Hailey Nguyen and
                  Hannah Korevaar and
                  Hu Xu and
                  Hugo Touvron and
                  Iliyan Zarov and
                  Imanol Arrieta Ibarra and
                  Isabel M. Kloumann and
                  Ishan Misra and
                  Ivan Evtimov and
                  Jade Copet and
                  Jaewon Lee and
                  Jan Geffert and
                  Jana Vranes and
                  Jason Park and
                  Jay Mahadeokar and
                  Jeet Shah and
                  Jelmer van der Linde and
                  Jennifer Billock and
                  Jenny Hong and
                  Jenya Lee and
                  Jeremy Fu and
                  Jianfeng Chi and
                  Jianyu Huang and
                  Jiawen Liu and
                  Jie Wang and
                  Jiecao Yu and
                  Joanna Bitton and
                  Joe Spisak and
                  Jongsoo Park and
                  Joseph Rocca and
                  Joshua Johnstun and
                  Joshua Saxe and
                  Junteng Jia and
                  Kalyan Vasuden Alwala and
                  Kartikeya Upasani and
                  Kate Plawiak and
                  Ke Li and
                  Kenneth Heafield and
                  Kevin Stone and
                  et al.},
  title        = {The Llama 3 Herd of Models},
  journal      = {CoRR},
  volume       = {abs/2407.21783},
  year         = {2024},
  url          = {https://doi.org/10.48550/arXiv.2407.21783},
  doi          = {10.48550/ARXIV.2407.21783},
  eprinttype    = {arXiv},
  eprint       = {2407.21783},
  bibsource    = {dblp computer science bibliography, https://dblp.org}
}

@article{DBLP:journals/corr/abs-2411-15124,
  author       = {Nathan Lambert and
                  Jacob Morrison and
                  Valentina Pyatkin and
                  Shengyi Huang and
                  Hamish Ivison and
                  Faeze Brahman and
                  Lester James V. Miranda and
                  Alisa Liu and
                  Nouha Dziri and
                  Shane Lyu and
                  Yuling Gu and
                  Saumya Malik and
                  Victoria Graf and
                  Jena D. Hwang and
                  Jiangjiang Yang and
                  Ronan Le Bras and
                  Oyvind Tafjord and
                  Chris Wilhelm and
                  Luca Soldaini and
                  Noah A. Smith and
                  Yizhong Wang and
                  Pradeep Dasigi and
                  Hannaneh Hajishirzi},
  title        = {T{\"{U}}LU 3: Pushing Frontiers in Open Language Model Post-Training},
  journal      = {CoRR},
  volume       = {abs/2411.15124},
  year         = {2024},
  url          = {https://doi.org/10.48550/arXiv.2411.15124},
  doi          = {10.48550/ARXIV.2411.15124},
  eprinttype    = {arXiv},
  eprint       = {2411.15124},
  bibsource    = {dblp computer science bibliography, https://dblp.org}
}

@article{DBLP:journals/corr/abs-2310-01405,
  author       = {Andy Zou and
                  Long Phan and
                  Sarah Li Chen and
                  James Campbell and
                  Phillip Guo and
                  Richard Ren and
                  Alexander Pan and
                  Xuwang Yin and
                  Mantas Mazeika and
                  Ann{-}Kathrin Dombrowski and
                  Shashwat Goel and
                  Nathaniel Li and
                  Michael J. Byun and
                  Zifan Wang and
                  Alex Mallen and
                  Steven Basart and
                  Sanmi Koyejo and
                  Dawn Song and
                  Matt Fredrikson and
                  J. Zico Kolter and
                  Dan Hendrycks},
  title        = {Representation Engineering: {A} Top-Down Approach to {AI} Transparency},
  journal      = {CoRR},
  volume       = {abs/2310.01405},
  year         = {2023},
  url          = {https://doi.org/10.48550/arXiv.2310.01405},
  doi          = {10.48550/ARXIV.2310.01405},
  eprinttype    = {arXiv},
  eprint       = {2310.01405},
  bibsource    = {dblp computer science bibliography, https://dblp.org}
}

@inproceedings{DBLP:conf/iclr/LightmanKBEBLLS24,
  author       = {Hunter Lightman and
                  Vineet Kosaraju and
                  Yuri Burda and
                  Harrison Edwards and
                  Bowen Baker and
                  Teddy Lee and
                  Jan Leike and
                  John Schulman and
                  Ilya Sutskever and
                  Karl Cobbe},
  title        = {Let's Verify Step by Step},
  booktitle    = {The Twelfth International Conference on Learning Representations,
                  {ICLR} 2024, Vienna, Austria, May 7-11, 2024},
  publisher    = {OpenReview.net},
  year         = {2024},
  url          = {https://openreview.net/forum?id=v8L0pN6EOi},
  bibsource    = {dblp computer science bibliography, https://dblp.org}
}

@article{DBLP:journals/corr/abs-2107-03374,
  author       = {Mark Chen and
                  Jerry Tworek and
                  Heewoo Jun and
                  Qiming Yuan and
                  Henrique Pond{\'{e}} de Oliveira Pinto and
                  Jared Kaplan and
                  Harri Edwards and
                  Yuri Burda and
                  Nicholas Joseph and
                  Greg Brockman and
                  Alex Ray and
                  Raul Puri and
                  Gretchen Krueger and
                  Michael Petrov and
                  Heidy Khlaaf and
                  Girish Sastry and
                  Pamela Mishkin and
                  Brooke Chan and
                  Scott Gray and
                  Nick Ryder and
                  Mikhail Pavlov and
                  Alethea Power and
                  Lukasz Kaiser and
                  Mohammad Bavarian and
                  Clemens Winter and
                  Philippe Tillet and
                  Felipe Petroski Such and
                  Dave Cummings and
                  Matthias Plappert and
                  Fotios Chantzis and
                  Elizabeth Barnes and
                  Ariel Herbert{-}Voss and
                  William Hebgen Guss and
                  Alex Nichol and
                  Alex Paino and
                  Nikolas Tezak and
                  Jie Tang and
                  Igor Babuschkin and
                  Suchir Balaji and
                  Shantanu Jain and
                  William Saunders and
                  Christopher Hesse and
                  Andrew N. Carr and
                  Jan Leike and
                  Joshua Achiam and
                  Vedant Misra and
                  Evan Morikawa and
                  Alec Radford and
                  Matthew Knight and
                  Miles Brundage and
                  Mira Murati and
                  Katie Mayer and
                  Peter Welinder and
                  Bob McGrew and
                  Dario Amodei and
                  Sam McCandlish and
                  Ilya Sutskever and
                  Wojciech Zaremba},
  title        = {Evaluating Large Language Models Trained on Code},
  journal      = {CoRR},
  volume       = {abs/2107.03374},
  year         = {2021},
  url          = {https://arxiv.org/abs/2107.03374},
  eprinttype    = {arXiv},
  eprint       = {2107.03374},
  bibsource    = {dblp computer science bibliography, https://dblp.org}
}

@inproceedings{DBLP:conf/nips/HendrycksBKABTS21,
  author       = {Dan Hendrycks and
                  Collin Burns and
                  Saurav Kadavath and
                  Akul Arora and
                  Steven Basart and
                  Eric Tang and
                  Dawn Song and
                  Jacob Steinhardt},
  editor       = {Joaquin Vanschoren and
                  Sai{-}Kit Yeung},
  title        = {Measuring Mathematical Problem Solving With the {MATH} Dataset},
  booktitle    = {Proceedings of the Neural Information Processing Systems Track on
                  Datasets and Benchmarks 1, NeurIPS Datasets and Benchmarks 2021, December
                  2021, virtual},
  year         = {2021},
  url          = {https://datasets-benchmarks-proceedings.neurips.cc/paper/2021/hash/be83ab3ecd0db773eb2dc1b0a17836a1-Abstract-round2.html},
  bibsource    = {dblp computer science bibliography, https://dblp.org}
}

@inproceedings{DBLP:conf/iclr/HuSWALWWC22,
  author       = {Edward J. Hu and
                  Yelong Shen and
                  Phillip Wallis and
                  Zeyuan Allen{-}Zhu and
                  Yuanzhi Li and
                  Shean Wang and
                  Lu Wang and
                  Weizhu Chen},
  title        = {LoRA: Low-Rank Adaptation of Large Language Models},
  booktitle    = {The Tenth International Conference on Learning Representations, {ICLR}
                  2022, Virtual Event, April 25-29, 2022},
  publisher    = {OpenReview.net},
  year         = {2022},
  url          = {https://openreview.net/forum?id=nZeVKeeFYf9},
  bibsource    = {dblp computer science bibliography, https://dblp.org}
}

@inproceedings{DBLP:conf/iclr/AgarwalVZSGGB24,
  author       = {Rishabh Agarwal and
                  Nino Vieillard and
                  Yongchao Zhou and
                  Piotr Stanczyk and
                  Sabela Ramos Garea and
                  Matthieu Geist and
                  Olivier Bachem},
  title        = {On-Policy Distillation of Language Models: Learning from Self-Generated
                  Mistakes},
  booktitle    = {The Twelfth International Conference on Learning Representations,
                  {ICLR} 2024, Vienna, Austria, May 7-11, 2024},
  publisher    = {OpenReview.net},
  year         = {2024},
  url          = {https://openreview.net/forum?id=3zKtaqxLhW},
  bibsource    = {dblp computer science bibliography, https://dblp.org}
}

@article{lu2025onpolicydistillation,
  author = {Kevin Lu and Thinking Machines Lab},
  title = {On-Policy Distillation},
  journal = {Thinking Machines Lab: Connectionism},
  year = {2025},
  url = {https://thinkingmachines.ai/blog/on-policy-distillation},
  doi = {10.64434/tml.20251026},
}

@inproceedings{DBLP:conf/nips/RafailovSMMEF23,
  author       = {Rafael Rafailov and
                  Archit Sharma and
                  Eric Mitchell and
                  Christopher D. Manning and
                  Stefano Ermon and
                  Chelsea Finn},
  editor       = {Alice Oh and
                  Tristan Naumann and
                  Amir Globerson and
                  Kate Saenko and
                  Moritz Hardt and
                  Sergey Levine},
  title        = {Direct Preference Optimization: Your Language Model is Secretly a
                  Reward Model},
  booktitle    = {Advances in Neural Information Processing Systems 36: Annual Conference
                  on Neural Information Processing Systems 2023, NeurIPS 2023, New Orleans,
                  LA, USA, December 10 - 16, 2023},
  year         = {2023},
  url          = {http://papers.nips.cc/paper\_files/paper/2023/hash/a85b405ed65c6477a4fe8302b5e06ce7-Abstract-Conference.html},
  bibsource    = {dblp computer science bibliography, https://dblp.org}
}

@article{luo2025deepscaler,
  title={Deepscaler: Surpassing o1-preview with a 1.5 b model by scaling rl},
  author={Luo, Michael and Tan, Sijun and Wong, Justin and Shi, Xiaoxiang and Tang, William Y and Roongta, Manan and Cai, Colin and Luo, Jeffrey and Zhang, Tianjun and Li, Li Erran and others},
  journal={Notion Blog},
  year={2025}
}

@article{liu2025understanding,
  title={Understanding r1-zero-like training: A critical perspective},
  author={Liu, Zichen and Chen, Changyu and Li, Wenjun and Qi, Penghui and Pang, Tianyu and Du, Chao and Lee, Wee Sun and Lin, Min},
  journal={arXiv preprint arXiv:2503.20783},
  year={2025}
}

@article{wang2025reinforcement,
  title={Reinforcement learning for reasoning in large language models with one training example},
  author={Wang, Yiping and Yang, Qing and Zeng, Zhiyuan and Ren, Liliang and Liu, Lucas and Peng, Baolin and Cheng, Hao and He, Xuehai and Wang, Kuan and Gao, Jianfeng and others},
  journal={arXiv preprint arXiv:2504.20571},
  year={2025}
}

@inproceedings{DBLP:conf/nips/FanL0TQCC24,
  author       = {Chenghao Fan and
                  Zhenyi Lu and
                  Wei Wei and
                  Jie Tian and
                  Xiaoye Qu and
                  Dangyang Chen and
                  Yu Cheng},
  editor       = {Amir Globersons and
                  Lester Mackey and
                  Danielle Belgrave and
                  Angela Fan and
                  Ulrich Paquet and
                  Jakub M. Tomczak and
                  Cheng Zhang},
  title        = {On Giant's Shoulders: Effortless Weak to Strong by Dynamic Logits
                  Fusion},
  booktitle    = {Advances in Neural Information Processing Systems 38: Annual Conference
                  on Neural Information Processing Systems 2024, NeurIPS 2024, Vancouver,
                  BC, Canada, December 10 - 15, 2024},
  year         = {2024},
  url          = {http://papers.nips.cc/paper\_files/paper/2024/hash/34ec1286b2ccd4794c5ca4ad078b7150-Abstract-Conference.html},
  bibsource    = {dblp computer science bibliography, https://dblp.org}
}

@article{liutuning,
  author       = {Alisa Liu and
                  Xiaochuang Han and
                  Yizhong Wang and
                  Yulia Tsvetkov and
                  Yejin Choi and
                  Noah A. Smith},
  title        = {Tuning Language Models by Proxy},
  journal      = {CoRR},
  volume       = {abs/2401.08565},
  year         = {2024},
  url          = {https://doi.org/10.48550/arXiv.2401.08565},
  doi          = {10.48550/ARXIV.2401.08565},
  eprinttype    = {arXiv},
  eprint       = {2401.08565},
  bibsource    = {dblp computer science bibliography, https://dblp.org}
}

@inproceedings{mitchellemulator,
  author       = {Eric Mitchell and
                  Rafael Rafailov and
                  Archit Sharma and
                  Chelsea Finn and
                  Christopher D. Manning},
  title        = {An Emulator for Fine-tuning Large Language Models using Small Language
                  Models},
  booktitle    = {The Twelfth International Conference on Learning Representations,
                  {ICLR} 2024, Vienna, Austria, May 7-11, 2024},
  publisher    = {OpenReview.net},
  year         = {2024},
  url          = {https://openreview.net/forum?id=Eo7kv0sllr},
  bibsource    = {dblp computer science bibliography, https://dblp.org}
}

@article{DBLP:journals/corr/abs-2502-03387,
  author       = {Yixin Ye and
                  Zhen Huang and
                  Yang Xiao and
                  Ethan Chern and
                  Shijie Xia and
                  Pengfei Liu},
  title        = {{LIMO:} Less is More for Reasoning},
  journal      = {CoRR},
  volume       = {abs/2502.03387},
  year         = {2025},
  url          = {https://doi.org/10.48550/arXiv.2502.03387},
  doi          = {10.48550/ARXIV.2502.03387},
  eprinttype    = {arXiv},
  eprint       = {2502.03387},
  bibsource    = {dblp computer science bibliography, https://dblp.org}
}

@article{DBLP:journals/corr/abs-2311-12022,
  author       = {David Rein and
                  Betty Li Hou and
                  Asa Cooper Stickland and
                  Jackson Petty and
                  Richard Yuanzhe Pang and
                  Julien Dirani and
                  Julian Michael and
                  Samuel R. Bowman},
  title        = {{GPQA:} {A} Graduate-Level Google-Proof Q{\&}A Benchmark},
  journal      = {CoRR},
  volume       = {abs/2311.12022},
  year         = {2023},
  url          = {https://doi.org/10.48550/arXiv.2311.12022},
  doi          = {10.48550/ARXIV.2311.12022},
  eprinttype    = {arXiv},
  eprint       = {2311.12022},
  bibsource    = {dblp computer science bibliography, https://dblp.org}
}

@article{DBLP:journals/corr/abs-2308-01825,
  author       = {Zheng Yuan and
                  Hongyi Yuan and
                  Chengpeng Li and
                  Guanting Dong and
                  Chuanqi Tan and
                  Chang Zhou},
  title        = {Scaling Relationship on Learning Mathematical Reasoning with Large
                  Language Models},
  journal      = {CoRR},
  volume       = {abs/2308.01825},
  year         = {2023},
  url          = {https://doi.org/10.48550/arXiv.2308.01825},
  doi          = {10.48550/ARXIV.2308.01825},
  eprinttype    = {arXiv},
  eprint       = {2308.01825},
  bibsource    = {dblp computer science bibliography, https://dblp.org}
}

@inproceedings{DBLP:conf/nips/BrownMRSKDNSSAA20,
  author       = {Tom B. Brown and
                  Benjamin Mann and
                  Nick Ryder and
                  Melanie Subbiah and
                  Jared Kaplan and
                  Prafulla Dhariwal and
                  Arvind Neelakantan and
                  Pranav Shyam and
                  Girish Sastry and
                  Amanda Askell and
                  Sandhini Agarwal and
                  Ariel Herbert{-}Voss and
                  Gretchen Krueger and
                  Tom Henighan and
                  Rewon Child and
                  Aditya Ramesh and
                  Daniel M. Ziegler and
                  Jeffrey Wu and
                  Clemens Winter and
                  Christopher Hesse and
                  Mark Chen and
                  Eric Sigler and
                  Mateusz Litwin and
                  Scott Gray and
                  Benjamin Chess and
                  Jack Clark and
                  Christopher Berner and
                  Sam McCandlish and
                  Alec Radford and
                  Ilya Sutskever and
                  Dario Amodei},
  editor       = {Hugo Larochelle and
                  Marc'Aurelio Ranzato and
                  Raia Hadsell and
                  Maria{-}Florina Balcan and
                  Hsuan{-}Tien Lin},
  title        = {Language Models are Few-Shot Learners},
  booktitle    = {Advances in Neural Information Processing Systems 33: Annual Conference
                  on Neural Information Processing Systems 2020, NeurIPS 2020, December
                  6-12, 2020, virtual},
  year         = {2020},
  url          = {https://proceedings.neurips.cc/paper/2020/hash/1457c0d6bfcb4967418bfb8ac142f64a-Abstract.html},
  bibsource    = {dblp computer science bibliography, https://dblp.org}
}

@inproceedings{DBLP:conf/emnlp/MinLHALHZ22,
  author       = {Sewon Min and
                  Xinxi Lyu and
                  Ari Holtzman and
                  Mikel Artetxe and
                  Mike Lewis and
                  Hannaneh Hajishirzi and
                  Luke Zettlemoyer},
  editor       = {Yoav Goldberg and
                  Zornitsa Kozareva and
                  Yue Zhang},
  title        = {Rethinking the Role of Demonstrations: What Makes In-Context Learning
                  Work?},
  booktitle    = {Proceedings of the 2022 Conference on Empirical Methods in Natural
                  Language Processing, {EMNLP} 2022, Abu Dhabi, United Arab Emirates,
                  December 7-11, 2022},
  pages        = {11048--11064},
  publisher    = {Association for Computational Linguistics},
  year         = {2022},
  url          = {https://doi.org/10.18653/v1/2022.emnlp-main.759},
  doi          = {10.18653/V1/2022.EMNLP-MAIN.759},
  bibsource    = {dblp computer science bibliography, https://dblp.org}
}

@article{DBLP:journals/corr/abs-2402-03300,
  author       = {Zhihong Shao and
                  Peiyi Wang and
                  Qihao Zhu and
                  Runxin Xu and
                  Junxiao Song and
                  Mingchuan Zhang and
                  Y. K. Li and
                  Y. Wu and
                  Daya Guo},
  title        = {DeepSeekMath: Pushing the Limits of Mathematical Reasoning in Open
                  Language Models},
  journal      = {CoRR},
  volume       = {abs/2402.03300},
  year         = {2024},
  url          = {https://doi.org/10.48550/arXiv.2402.03300},
  doi          = {10.48550/ARXIV.2402.03300},
  eprinttype    = {arXiv},
  eprint       = {2402.03300},
  bibsource    = {dblp computer science bibliography, https://dblp.org}
}

@article{DBLP:journals/corr/SchulmanWDRK17,
  author       = {John Schulman and
                  Filip Wolski and
                  Prafulla Dhariwal and
                  Alec Radford and
                  Oleg Klimov},
  title        = {Proximal Policy Optimization Algorithms},
  journal      = {CoRR},
  volume       = {abs/1707.06347},
  year         = {2017},
  url          = {http://arxiv.org/abs/1707.06347},
  eprinttype    = {arXiv},
  eprint       = {1707.06347},
  bibsource    = {dblp computer science bibliography, https://dblp.org}
}

@inproceedings{DBLP:conf/emnlp/Dong0DZMLXX0C0S24,
  author       = {Qingxiu Dong and
                  Lei Li and
                  Damai Dai and
                  Ce Zheng and
                  Jingyuan Ma and
                  Rui Li and
                  Heming Xia and
                  Jingjing Xu and
                  Zhiyong Wu and
                  Baobao Chang and
                  Xu Sun and
                  Lei Li and
                  Zhifang Sui},
  editor       = {Yaser Al{-}Onaizan and
                  Mohit Bansal and
                  Yun{-}Nung Chen},
  title        = {A Survey on In-context Learning},
  booktitle    = {Proceedings of the 2024 Conference on Empirical Methods in Natural
                  Language Processing, {EMNLP} 2024, Miami, FL, USA, November 12-16,
                  2024},
  pages        = {1107--1128},
  publisher    = {Association for Computational Linguistics},
  year         = {2024},
  url          = {https://doi.org/10.18653/v1/2024.emnlp-main.64},
  doi          = {10.18653/V1/2024.EMNLP-MAIN.64},
  bibsource    = {dblp computer science bibliography, https://dblp.org}
}

@article{DBLP:journals/corr/abs-2501-19393,
  author       = {Niklas Muennighoff and
                  Zitong Yang and
                  Weijia Shi and
                  Xiang Lisa Li and
                  Li Fei{-}Fei and
                  Hannaneh Hajishirzi and
                  Luke Zettlemoyer and
                  Percy Liang and
                  Emmanuel J. Cand{\`{e}}s and
                  Tatsunori Hashimoto},
  title        = {s1: Simple test-time scaling},
  journal      = {CoRR},
  volume       = {abs/2501.19393},
  year         = {2025},
  url          = {https://doi.org/10.48550/arXiv.2501.19393},
  doi          = {10.48550/ARXIV.2501.19393},
  eprinttype    = {arXiv},
  eprint       = {2501.19393},
  bibsource    = {dblp computer science bibliography, https://dblp.org}
}

@article{DBLP:journals/corr/abs-2501-11284,
  author       = {Haotian Xu and
                  Xing Wu and
                  Weinong Wang and
                  Zhongzhi Li and
                  Da Zheng and
                  Boyuan Chen and
                  Yi Hu and
                  Shijia Kang and
                  Jiaming Ji and
                  Yingying Zhang and
                  Zhijiang Guo and
                  Yaodong Yang and
                  Muhan Zhang and
                  Debing Zhang},
  title        = {RedStar: Does Scaling Long-CoT Data Unlock Better Slow-Reasoning Systems?},
  journal      = {CoRR},
  volume       = {abs/2501.11284},
  year         = {2025},
  url          = {https://doi.org/10.48550/arXiv.2501.11284},
  doi          = {10.48550/ARXIV.2501.11284},
  eprinttype    = {arXiv},
  eprint       = {2501.11284},
  bibsource    = {dblp computer science bibliography, https://dblp.org}
}

@article{DBLP:journals/corr/abs-2503-14476,
  author       = {Qiying Yu and
                  Zheng Zhang and
                  Ruofei Zhu and
                  Yufeng Yuan and
                  Xiaochen Zuo and
                  Yu Yue and
                  Tiantian Fan and
                  Gaohong Liu and
                  Lingjun Liu and
                  Xin Liu and
                  Haibin Lin and
                  Zhiqi Lin and
                  Bole Ma and
                  Guangming Sheng and
                  Yuxuan Tong and
                  Chi Zhang and
                  Mofan Zhang and
                  Wang Zhang and
                  Hang Zhu and
                  Jinhua Zhu and
                  Jiaze Chen and
                  Jiangjie Chen and
                  Chengyi Wang and
                  Hongli Yu and
                  Weinan Dai and
                  Yuxuan Song and
                  Xiangpeng Wei and
                  Hao Zhou and
                  Jingjing Liu and
                  Wei{-}Ying Ma and
                  Ya{-}Qin Zhang and
                  Lin Yan and
                  Mu Qiao and
                  Yonghui Wu and
                  Mingxuan Wang},
  title        = {{DAPO:} An Open-Source {LLM} Reinforcement Learning System at Scale},
  journal      = {CoRR},
  volume       = {abs/2503.14476},
  year         = {2025},
  url          = {https://doi.org/10.48550/arXiv.2503.14476},
  doi          = {10.48550/ARXIV.2503.14476},
  eprinttype    = {arXiv},
  eprint       = {2503.14476},
  bibsource    = {dblp computer science bibliography, https://dblp.org}
}

@article{DBLP:journals/corr/abs-2503-01307,
  author       = {Kanishk Gandhi and
                  Ayush Chakravarthy and
                  Anikait Singh and
                  Nathan Lile and
                  Noah D. Goodman},
  title        = {Cognitive Behaviors that Enable Self-Improving Reasoners, or, Four
                  Habits of Highly Effective STaRs},
  journal      = {CoRR},
  volume       = {abs/2503.01307},
  year         = {2025},
  url          = {https://doi.org/10.48550/arXiv.2503.01307},
  doi          = {10.48550/ARXIV.2503.01307},
  eprinttype    = {arXiv},
  eprint       = {2503.01307},
  bibsource    = {dblp computer science bibliography, https://dblp.org}
}

@online{OpenAI2024LearningToReason,
  title={{Learning to reason with LLMs}},
  author={{OpenAI}},
  year={2024},
  month={September},
  day={12},
  url={https://openai.com/index/learning-to-reason-with-llms/},
  urldate={2025-05-23}
}

@article{DBLP:journals/corr/abs-2503-18892,
  author       = {Weihao Zeng and
                  Yuzhen Huang and
                  Qian Liu and
                  Wei Liu and
                  Keqing He and
                  Zejun Ma and
                  Junxian He},
  title        = {SimpleRL-Zoo: Investigating and Taming Zero Reinforcement Learning
                  for Open Base Models in the Wild},
  journal      = {CoRR},
  volume       = {abs/2503.18892},
  year         = {2025},
  url          = {https://doi.org/10.48550/arXiv.2503.18892},
  doi          = {10.48550/ARXIV.2503.18892},
  eprinttype    = {arXiv},
  eprint       = {2503.18892},
  bibsource    = {dblp computer science bibliography, https://dblp.org}
}

@article{DBLP:journals/corr/abs-2501-12948,
  author       = {DeepSeek{-}AI and
                  Daya Guo and
                  Dejian Yang and
                  Haowei Zhang and
                  Junxiao Song and
                  Ruoyu Zhang and
                  Runxin Xu and
                  Qihao Zhu and
                  Shirong Ma and
                  Peiyi Wang and
                  Xiao Bi and
                  Xiaokang Zhang and
                  Xingkai Yu and
                  Yu Wu and
                  Z. F. Wu and
                  Zhibin Gou and
                  Zhihong Shao and
                  Zhuoshu Li and
                  Ziyi Gao and
                  Aixin Liu and
                  Bing Xue and
                  Bingxuan Wang and
                  Bochao Wu and
                  Bei Feng and
                  Chengda Lu and
                  Chenggang Zhao and
                  Chengqi Deng and
                  Chenyu Zhang and
                  Chong Ruan and
                  Damai Dai and
                  Deli Chen and
                  Dongjie Ji and
                  Erhang Li and
                  Fangyun Lin and
                  Fucong Dai and
                  Fuli Luo and
                  Guangbo Hao and
                  Guanting Chen and
                  Guowei Li and
                  H. Zhang and
                  Han Bao and
                  Hanwei Xu and
                  Haocheng Wang and
                  Honghui Ding and
                  Huajian Xin and
                  Huazuo Gao and
                  Hui Qu and
                  Hui Li and
                  Jianzhong Guo and
                  Jiashi Li and
                  Jiawei Wang and
                  Jingchang Chen and
                  Jingyang Yuan and
                  Junjie Qiu and
                  Junlong Li and
                  J. L. Cai and
                  Jiaqi Ni and
                  Jian Liang and
                  Jin Chen and
                  Kai Dong and
                  Kai Hu and
                  Kaige Gao and
                  Kang Guan and
                  Kexin Huang and
                  Kuai Yu and
                  Lean Wang and
                  Lecong Zhang and
                  Liang Zhao and
                  Litong Wang and
                  Liyue Zhang and
                  Lei Xu and
                  Leyi Xia and
                  Mingchuan Zhang and
                  Minghua Zhang and
                  Minghui Tang and
                  Meng Li and
                  Miaojun Wang and
                  Mingming Li and
                  Ning Tian and
                  Panpan Huang and
                  Peng Zhang and
                  Qiancheng Wang and
                  Qinyu Chen and
                  Qiushi Du and
                  Ruiqi Ge and
                  Ruisong Zhang and
                  Ruizhe Pan and
                  Runji Wang and
                  R. J. Chen and
                  R. L. Jin and
                  Ruyi Chen and
                  Shanghao Lu and
                  Shangyan Zhou and
                  Shanhuang Chen and
                  Shengfeng Ye and
                  Shiyu Wang and
                  Shuiping Yu and
                  Shunfeng Zhou and
                  Shuting Pan and
                  S. S. Li},
  title        = {DeepSeek-R1: Incentivizing Reasoning Capability in LLMs via Reinforcement
                  Learning},
  journal      = {CoRR},
  volume       = {abs/2501.12948},
  year         = {2025},
  url          = {https://doi.org/10.48550/arXiv.2501.12948},
  doi          = {10.48550/ARXIV.2501.12948},
  eprinttype    = {arXiv},
  eprint       = {2501.12948},
  bibsource    = {dblp computer science bibliography, https://dblp.org}
}

@article{DBLP:journals/tmlr/HuangZ0M0PC25,
  author       = {James Y. Huang and
                  Wenxuan Zhou and
                  Fei Wang and
                  Fred Morstatter and
                  Sheng Zhang and
                  Hoifung Poon and
                  Muhao Chen},
  title        = {Offset Unlearning for Large Language Models},
  journal      = {Trans. Mach. Learn. Res.},
  volume       = {2025},
  year         = {2025},
  url          = {https://openreview.net/forum?id=A4RLpHPXCu},
  bibsource    = {dblp computer science bibliography, https://dblp.org}
}

@article{DBLP:journals/corr/abs-2401-17256,
  author       = {Xuandong Zhao and
                  Xianjun Yang and
                  Tianyu Pang and
                  Chao Du and
                  Lei Li and
                  Yu{-}Xiang Wang and
                  William Yang Wang},
  title        = {Weak-to-Strong Jailbreaking on Large Language Models},
  journal      = {CoRR},
  volume       = {abs/2401.17256},
  year         = {2024},
  url          = {https://doi.org/10.48550/arXiv.2401.17256},
  doi          = {10.48550/ARXIV.2401.17256},
  eprinttype    = {arXiv},
  eprint       = {2401.17256},
  bibsource    = {dblp computer science bibliography, https://dblp.org}
}

@inproceedings{liu2021dexperts,
  author       = {Alisa Liu and
                  Maarten Sap and
                  Ximing Lu and
                  Swabha Swayamdipta and
                  Chandra Bhagavatula and
                  Noah A. Smith and
                  Yejin Choi},
  editor       = {Chengqing Zong and
                  Fei Xia and
                  Wenjie Li and
                  Roberto Navigli},
  title        = {DExperts: Decoding-Time Controlled Text Generation with Experts and
                  Anti-Experts},
  booktitle    = {Proceedings of the 59th Annual Meeting of the Association for Computational
                  Linguistics and the 11th International Joint Conference on Natural
                  Language Processing, {ACL/IJCNLP} 2021, (Volume 1: Long Papers), Virtual
                  Event, August 1-6, 2021},
  pages        = {6691--6706},
  publisher    = {Association for Computational Linguistics},
  year         = {2021},
  url          = {https://doi.org/10.18653/v1/2021.acl-long.522},
  doi          = {10.18653/V1/2021.ACL-LONG.522},
  bibsource    = {dblp computer science bibliography, https://dblp.org}
}

@inproceedings{DBLP:conf/naacl/ShiHLTZY24,
  author       = {Weijia Shi and
                  Xiaochuang Han and
                  Mike Lewis and
                  Yulia Tsvetkov and
                  Luke Zettlemoyer and
                  Wen{-}tau Yih},
  editor       = {Kevin Duh and
                  Helena G{\'{o}}mez{-}Adorno and
                  Steven Bethard},
  title        = {Trusting Your Evidence: Hallucinate Less with Context-aware Decoding},
  booktitle    = {Proceedings of the 2024 Conference of the North American Chapter of
                  the Association for Computational Linguistics: Human Language Technologies:
                  Short Papers, {NAACL} 2024, Mexico City, Mexico, June 16-21, 2024},
  pages        = {783--791},
  publisher    = {Association for Computational Linguistics},
  year         = {2024},
  url          = {https://doi.org/10.18653/v1/2024.naacl-short.69},
  doi          = {10.18653/V1/2024.NAACL-SHORT.69},
  bibsource    = {dblp computer science bibliography, https://dblp.org}
}

@inproceedings{DBLP:conf/emnlp/OrmazabalAA23,
  author       = {Aitor Ormazabal and
                  Mikel Artetxe and
                  Eneko Agirre},
  editor       = {Houda Bouamor and
                  Juan Pino and
                  Kalika Bali},
  title        = {CombLM: Adapting Black-Box Language Models through Small Fine-Tuned
                  Models},
  booktitle    = {Proceedings of the 2023 Conference on Empirical Methods in Natural
                  Language Processing, {EMNLP} 2023, Singapore, December 6-10, 2023},
  pages        = {2961--2974},
  publisher    = {Association for Computational Linguistics},
  year         = {2023},
  url          = {https://doi.org/10.18653/v1/2023.emnlp-main.180},
  doi          = {10.18653/V1/2023.EMNLP-MAIN.180},
  bibsource    = {dblp computer science bibliography, https://dblp.org}
}

@article{zhao2025activation,
  title={Activation Control for Efficiently Eliciting Long Chain-of-thought Ability of Language Models},
  author={Zhao, Zekai and Liu, Qi and Zhou, Kun and Liu, Zihan and Shao, Yifei and Hu, Zhiting and Huang, Biwei},
  journal={arXiv preprint arXiv:2505.17697},
  year={2025}
}

@article{tang2025unlocking,
  title={Unlocking General Long Chain-of-Thought Reasoning Capabilities of Large Language Models via Representation Engineering},
  author={Tang, Xinyu and Wang, Xiaolei and Lv, Zhihao and Min, Yingqian and Zhao, Wayne Xin and Hu, Binbin and Liu, Ziqi and Zhang, Zhiqiang},
  journal={arXiv preprint arXiv:2503.11314},
  year={2025}
}

@article{li2025llms,
  author       = {Dacheng Li and
                  Shiyi Cao and
                  Tyler Griggs and
                  Shu Liu and
                  Xiangxi Mo and
                  Eric Tang and
                  Sumanth Hegde and
                  Kourosh Hakhamaneshi and
                  Shishir G. Patil and
                  Matei Zaharia and
                  Joseph E. Gonzalez and
                  Ion Stoica},
  title        = {LLMs Can Easily Learn to Reason from Demonstrations Structure, not
                  content, is what matters!},
  journal      = {CoRR},
  volume       = {abs/2502.07374},
  year         = {2025},
  url          = {https://doi.org/10.48550/arXiv.2502.07374},
  doi          = {10.48550/ARXIV.2502.07374},
  eprinttype    = {arXiv},
  eprint       = {2502.07374},
  bibsource    = {dblp computer science bibliography, https://dblp.org}
}

@article{dang2025reinforcement,
  title={Reinforcement Learning for Reasoning in Small LLMs: What Works and What Doesn't},
  author={Dang, Quy-Anh and Ngo, Chris},
  journal={arXiv preprint arXiv:2503.16219},
  year={2025}
}

@misc{deepscaler2025,
  title={DeepScaleR: Surpassing O1-Preview with a 1.5B Model by Scaling RL},
  author={Michael Luo and Sijun Tan and Justin Wong and Xiaoxiang Shi and William Y. Tang and Manan Roongta and Colin Cai and Jeffrey Luo and Li Erran Li and Raluca Ada Popa and Ion Stoica},
  year={2025},
  howpublished={\href{https://pretty-radio-b75.notion.site/DeepScaleR-Surpassing-O1-Preview-with-a-1-5B-Model-by-Scaling-RL-19681902c1468005bed8ca303013a4e2}{Notion Blog}},
  note={Accessed 2025}
}

@article{DBLP:journals/corr/abs-2409-12122,
  author       = {An Yang and
                  Beichen Zhang and
                  Binyuan Hui and
                  Bofei Gao and
                  Bowen Yu and
                  Chengpeng Li and
                  Dayiheng Liu and
                  Jianhong Tu and
                  Jingren Zhou and
                  Junyang Lin and
                  Keming Lu and
                  Mingfeng Xue and
                  Runji Lin and
                  Tianyu Liu and
                  Xingzhang Ren and
                  Zhenru Zhang},
  title        = {Qwen2.5-Math Technical Report: Toward Mathematical Expert Model via
                  Self-Improvement},
  journal      = {CoRR},
  volume       = {abs/2409.12122},
  year         = {2024},
  url          = {https://doi.org/10.48550/arXiv.2409.12122},
  doi          = {10.48550/ARXIV.2409.12122},
  eprinttype    = {arXiv},
  eprint       = {2409.12122},
  bibsource    = {dblp computer science bibliography, https://dblp.org}
}

@article{DBLP:journals/corr/abs-2412-15115,
  author       = {An Yang and
                  Baosong Yang and
                  Beichen Zhang and
                  Binyuan Hui and
                  Bo Zheng and
                  Bowen Yu and
                  Chengyuan Li and
                  Dayiheng Liu and
                  Fei Huang and
                  Haoran Wei and
                  Huan Lin and
                  Jian Yang and
                  Jianhong Tu and
                  Jianwei Zhang and
                  Jianxin Yang and
                  Jiaxi Yang and
                  Jingren Zhou and
                  Junyang Lin and
                  Kai Dang and
                  Keming Lu and
                  Keqin Bao and
                  Kexin Yang and
                  Le Yu and
                  Mei Li and
                  Mingfeng Xue and
                  Pei Zhang and
                  Qin Zhu and
                  Rui Men and
                  Runji Lin and
                  Tianhao Li and
                  Tingyu Xia and
                  Xingzhang Ren and
                  Xuancheng Ren and
                  Yang Fan and
                  Yang Su and
                  Yichang Zhang and
                  Yu Wan and
                  Yuqiong Liu and
                  Zeyu Cui and
                  Zhenru Zhang and
                  Zihan Qiu},
  title        = {Qwen2.5 Technical Report},
  journal      = {CoRR},
  volume       = {abs/2412.15115},
  year         = {2024},
  url          = {https://doi.org/10.48550/arXiv.2412.15115},
  doi          = {10.48550/ARXIV.2412.15115},
  eprinttype    = {arXiv},
  eprint       = {2412.15115},
  bibsource    = {dblp computer science bibliography, https://dblp.org}
}

@misc{openai_logitbias,
  title        = {Using logit bias to alter token probability with the OpenAI API},
  author       = {{OpenAI}},
  year         = {2025},
  url = {https://help.openai.com/en/articles/5247780-using-logit-bias-to-alter-token-probability-with-the-openai-api}}

@article{DBLP:journals/corr/abs-2409-00088,
  author       = {Jiajun Xu and
                  Zhiyuan Li and
                  Wei Chen and
                  Qun Wang and
                  Xin Gao and
                  Qi Cai and
                  Ziyuan Ling},
  title        = {On-Device Language Models: {A} Comprehensive Review},
  journal      = {CoRR},
  volume       = {abs/2409.00088},
  year         = {2024},
  url          = {https://doi.org/10.48550/arXiv.2409.00088},
  doi          = {10.48550/ARXIV.2409.00088},
  eprinttype    = {arXiv},
  eprint       = {2409.00088},
  bibsource    = {dblp computer science bibliography, https://dblp.org}
}

@inproceedings{DBLP:conf/nips/DettmersPHZ23,
  author       = {Tim Dettmers and
                  Artidoro Pagnoni and
                  Ari Holtzman and
                  Luke Zettlemoyer},
  editor       = {Alice Oh and
                  Tristan Naumann and
                  Amir Globerson and
                  Kate Saenko and
                  Moritz Hardt and
                  Sergey Levine},
  title        = {QLoRA: Efficient Finetuning of Quantized LLMs},
  booktitle    = {Advances in Neural Information Processing Systems 36: Annual Conference
                  on Neural Information Processing Systems 2023, NeurIPS 2023, New Orleans,
                  LA, USA, December 10 - 16, 2023},
  year         = {2023},
  url          = {http://papers.nips.cc/paper\_files/paper/2023/hash/1feb87871436031bdc0f2beaa62a049b-Abstract-Conference.html},
  bibsource    = {dblp computer science bibliography, https://dblp.org}
}

@inproceedings{DBLP:conf/aistats/McMahanMRHA17,
  author       = {Brendan McMahan and
                  Eider Moore and
                  Daniel Ramage and
                  Seth Hampson and
                  Blaise Ag{\"{u}}era y Arcas},
  editor       = {Aarti Singh and
                  Xiaojin (Jerry) Zhu},
  title        = {Communication-Efficient Learning of Deep Networks from Decentralized
                  Data},
  booktitle    = {Proceedings of the 20th International Conference on Artificial Intelligence
                  and Statistics, {AISTATS} 2017, 20-22 April 2017, Fort Lauderdale,
                  FL, {USA}},
  series       = {Proceedings of Machine Learning Research},
  volume       = {54},
  pages        = {1273--1282},
  publisher    = {{PMLR}},
  year         = {2017},
  url          = {http://proceedings.mlr.press/v54/mcmahan17a.html},
  bibsource    = {dblp computer science bibliography, https://dblp.org}
}

@software{360-llama-factory,
  author = {Haosheng Zou and  Xiaowei Lv and Shousheng Jia and Xiangzheng Zhang},
  title = {360-LLaMA-Factory},
  url = {https://github.com/Qihoo360/360-LLaMA-Factory},
  year = {2024}
}

@inproceedings{llamafactory,
  title={LlamaFactory: Unified Efficient Fine-Tuning of 100+ Language Models},
  author={Yaowei Zheng and Richong Zhang and Junhao Zhang and Yanhan Ye and Zheyan Luo and Zhangchi Feng and Yongqiang Ma},
  booktitle={Proceedings of the 62nd Annual Meeting of the Association for Computational Linguistics (Volume 3: System Demonstrations)},
  address={Bangkok, Thailand},
  publisher={Association for Computational Linguistics},
  year={2024},
  url={http://arxiv.org/abs/2403.13372}
}
\clearpage
\appendix


\section{Theoretical Grounding}
\label{app:theory}

We provide a clearer mathematical interpretation for why our logit arithmetic formulation behaves approximately linearly across tasks. 

Consider a pretrained model ($S_0$) and its post-trained long-CoT variant ($S$), which maximize an RL objective with a KL divergence penalty: 

$$
\max_{P(\cdot|x)} \mathbb{E}_{y\sim P} [r(x,y)] \ - \beta \  KL(P(\cdot|x) \| S_0(\cdot|x)).
$$

This is known to have a closed-form solution~\citep{DBLP:conf/nips/RafailovSMMEF23}:

$$
P^*(y|x)
=\frac{1}{Z(x)} S_0(y|x)\exp\left(\tfrac{1}{\beta}r(x,y)\right).
$$

Rearranging gives an implicit reward function for any fine-tuned model $S$:

$$
r(x,y)=\beta \log\frac{S(y|x)}{S_0(y|x)}.
$$

Thus, the log-likelihood ratio between a trained and untrained model encodes behavioral shift introduced by long-CoT training.

Now, in ThinkLogit we combine the target model $L$ with this reward signal via $\tilde{\ell} = \ell_L + \alpha \cdot (\ell_S - \ell_{S_0})$.

Exponentiating both sides gives:

$$
\tilde{P}(y|x)
\propto L(y|x)
\left(\frac{S(y|x)}{S_0(y|x)}\right)^{\alpha},
$$

which corresponds to adjusting the target distribution according to the guider’s implicit reward. In this view, the apparent linearity of logit arithmetic follows naturally from the multiplicative form of the KL-regularized optimum and the fact that logits are log-probabilities, rather than from an ad-hoc linearity assumption.

\section{Training Details}
\label{subsec:training-details}

\paragraph{Environment.}
All experiments were conducted using NVIDIA A40/L40S
GPUs with 48GB memory. The software environment was configured as follows:
\begin{itemize}[itemsep=2pt, topsep=2pt, parsep=0pt, partopsep=0pt]
  \item \texttt{360-LLaMA-Factory}~\citep{360-llama-factory} (A long CoT adapted version of \texttt{LLaMA-Factory}  0.9.1~\citep{llamafactory})
  \item \texttt{torch} 2.7.0
  \item \texttt{transformers} 4.51.3
  \item \texttt{accelerate} 1.0.1
  \item \texttt{datasets} 3.1.0
  \item \texttt{trl} 0.9.6
  \item \texttt{peft} 0.12.0
  \item \texttt{deepspeed} 0.14.4
\end{itemize}

\paragraph{LoRA Configuration.}
We applied LoRA~\citep{DBLP:conf/iclr/HuSWALWWC22} for parameter-efficient fine-tuning of the guider model:
\begin{itemize}[itemsep=2pt, topsep=2pt, parsep=0pt, partopsep=0pt]
  \item Rank: 64
  \item $\alpha_{\text{LoRA}}$: 128
  \item Target modules: \texttt{q\_proj}, \texttt{k\_proj}, \texttt{v\_proj}, \texttt{o\_proj}
  \item Bias: None
\end{itemize}

\paragraph{DPO Training.}
For preference optimization with DPO, we used the following settings:
\begin{itemize}[itemsep=2pt, topsep=2pt, parsep=0pt, partopsep=0pt]
  \item Batch size: 32 (4 GPUs * 8 Gradient Accumulation)
  \item Epoch: 1
  \item Learning rate: 5e-6
  \item Optimizer: AdamW
  \item Learning rate scheduler: cosine with warmup
  \item Warmup ratio: 0.1
  \item $\beta$ (reward scaling): 0.1
  \item Cutoff length: 8192
\end{itemize}

\paragraph{Preference Data Statistics.}
To construct the preference data, we sampled 5 completions for each question from both the guider ($S$) and target ($L$) models. Each completion was checked for final-answer correctness against the gold label. The target model $L$ yielded 12,412 correct ($y^{L\checkmark}$) and 16,448 incorrect ($y^{L\times}$) completions, whereas the guider $S$ produced 18,651 correct ($y^{S\checkmark}$) and 10,209 incorrect ($y^{S\times}$) completions. Forming the Cartesian product for each question resulted in 11,974 Type-1 preference pairs $\bigl(y^{L\checkmark}, y^{S\times}\bigr)$ and 43,209 Type-2 pairs $\bigl(y^{S\checkmark}, y^{L\times}\bigr)$, for a total pool of 55,183 pairs.

\section{Impact of Guider Model Size}
\label{app:guider_scaling}

A key hypothesis of \textsc{ThinkLogit} is that the quality of the elicited reasoning depends on the strength of the guidance signal provided by the small model. In our main experiments, we utilized a 1.5B parameter guider for efficiency. To investigate whether our approach benefits from scaling the guider's capacity, we replace the 1.5B guider with a larger 7B reasoning model (R1-Distill-Qwen-7B), while keeping the target model (Qwen2.5-32B) fixed.

Table~\ref{tab:guider_scaling} presents the results on AIME 2024 and AMC 23. We observe a strong positive correlation between guider size and performance gains. While the 1.5B guider yields reasonable improvements (+7.9\% absolute on AIME 2024), scaling to the 7B guider nearly triples this gain to +21.2\%. Similarly, on AMC 23, the improvement from base model jumps from +5.0\% to +12.7\% absolute.

These results demonstrate that \mybasemethod is not merely transferring a generic ``reasoning style'' (e.g., token verbosity), but effectively transferring the specific reasoning capabilities of the guider. Consequently, as small reasoning models continue to improve in quality and efficiency, our inference-time method can directly leverage these advances to further boost large frozen models.

\begin{table}
\centering
\setlength{\tabcolsep}{3pt}
\small
\begin{tabular}{lll}
\toprule
Model  & \makecell[l]{AIME\\2024} & \makecell[l]{AMC\\23} \\
\midrule
\textbf{\texttt{(Target)}} Qwen2.5-32B   & 14.6 &  57.2 \\
\textbf{\texttt{Target}} \colorbox{modelblue}{\,+ \mybasemethod \textbf{1.5B}\,}   & 22.5\mydelta{7.9} & 62.2\mydelta{5.0} \\
\textbf{\texttt{Target}} \colorbox{modelblue}{\,+ \mybasemethod \textbf{7B}\,}   & 35.8\mydelta{21.2} & 69.9\mydelta{12.7} \\
\bottomrule
\end{tabular}
\caption{
Scaling properties of the guider model. Avg@8 performance improves substantially when replacing the 1.5B guider with a 7B model, indicating that \mybasemethod effectively leverages increased capacity in the guider model.
}
\label{tab:guider_scaling}
\end{table}

\begin{table*}
\centering
\begin{tabular}{ll}
\toprule
\textbf{Behavior} & \textbf{Keywords} \\\midrule
Branching out & alternatively, another, try, suppose, consider, different, assume, also, option \\ 
Backtracking & however, but, mistake, error, contradiction, wrong, revisit, actually, again, flawed \\ 
Self-verification & check, verify, confirm, satisfy, plug, back, substitute, ensure, validate, test \\ \bottomrule
\end{tabular}
\caption{Keywords identifying three types of reasoning behaviors: Branching out, Backtracking, and Self-verification.}
\label{tab:keywords}
\end{table*}

\begin{figure*}
  \centering
  \includegraphics[width=\linewidth]{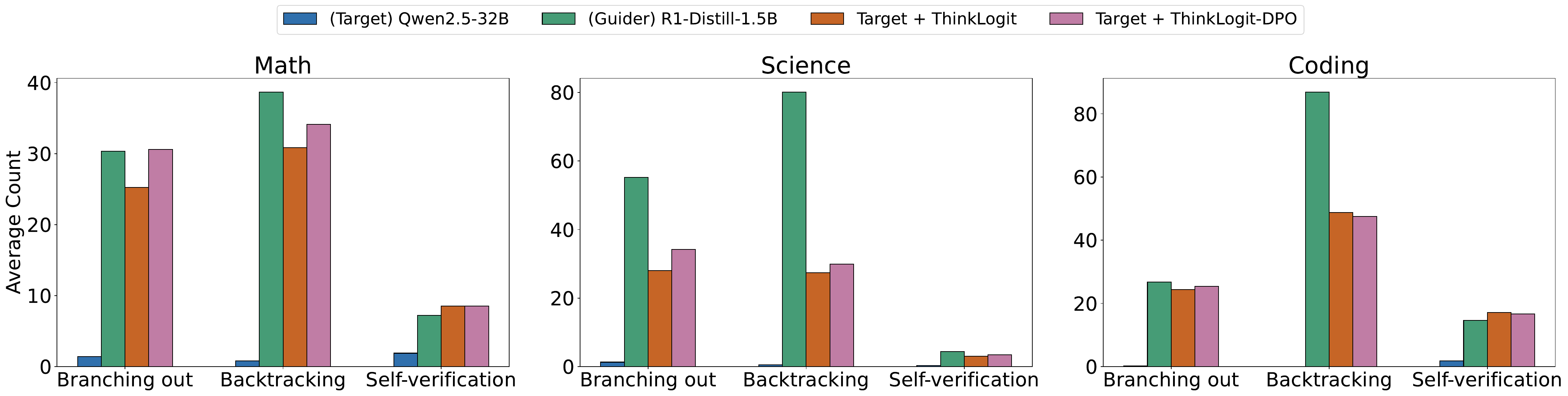}  
\caption{Domain-specific breakdown of reasoning behaviors across Math, Science, and Coding tasks. The elicited behaviors are robust across diverse domains.}
  \label{fig:reasoning_behaviour_combined}
\end{figure*}

\section{Domain-Specific Analysis on Reasoning Behaviors}
\label{app:behavior}

We further break down the reasoning behaviors defined in Section~\ref{sec:behavior} by domain in Figure~\ref{fig:reasoning_behaviour_combined}. The results show consistent trends across Math (AIME2024, AIME2025, AMC23, MATH Level 5), Science (GPQA Diamond), and Coding (LiveCodeBench) tasks. This detailed domain-specific breakdown confirms that the elicited reasoning behaviors are robust across different types of complex problem-solving scenarios and are not limited to a single domain.

\section{Case Study}
\label{app:case_study}

To determine whether \mybasemethod elicits meaningful thought processes or simply inflates output length, we compare the guider, target, and \mybasemethod outputs on a selected question from the MATH-500 dataset (Figure \ref{fig:ex}). The analysis demonstrates that \textsc{ThinkLogit} generates \textit{informative reasoning behaviors}. In contrast to existing training-free methods~\citep{DBLP:journals/corr/abs-2501-19393} that often artificially lengthen traces with repetitive filler, the additional tokens here drive functional self-correction, successfully redirecting an erroneous trajectory toward the true solution.

\section{Qualitative Analysis of Failure Modes}
\label{app:failure_modes}

While \mybasemethod demonstrates strong capabilities in eliciting long chain-of-thought reasoning without training, we observe specific failure modes where the interaction between the guider's impulse to lengthen generation and the target's knowledge boundaries leads to reasoning collapse. We categorize these failures into three distinct types.

\subsection{\mybasemethod struggles to form higher-level abstractions}
In this failure mode, the model successfully initiates a complex reasoning path and correctly sets up the problem constraints. However, when it lacks a specific high-level mathematical insight (e.g., recognizing an optimization boundary or an envelope), the guidance strength ($\alpha$) prevents the model from stopping or guessing. Instead, the model enters a recursive loop, re-deriving the same intermediate conditions using varied phrasing until the context window is exhausted.

\paragraph{Question:} \textit{Let $O(0,0), A(\tfrac{1}{2}, 0),$ and $B(0, \tfrac{\sqrt{3}}{2})$ be points in the coordinate plane. Let $\mathcal{F}$ be the family of segments $\overline{PQ}$ of unit length lying in the first quadrant with $P$ on the $x$-axis and $Q$ on the $y$-axis. There is a unique point $C$ on $\overline{AB}$, distinct from $A$ and $B$, that does not belong to any segment from $\mathcal{F}$ other than $\overline{AB}$. Then $OC^2 = \tfrac{p}{q}$, where $p$ and $q$ are relatively prime positive integers. Find $p + q$.}

\paragraph{Answer:} 23.

\paragraph{Failure Analysis:} The model correctly parameterizes the family of lines $y = (-q/p)x + q$ subject to $p^2+q^2=1$. However, it fails to recognize that the ``unique point'' corresponds to the point of tangency with the family's envelope (an astroid). Lacking this conceptual leap, the model attempts to prove the uniqueness condition algebraically for every possible parameter $p, q$.

\begin{tracebox}
    ``Therefore, I need to characterize all points $C$ on $AB$ such that there exists no other $PQ$ in $\mathcal{F}$ with $PQ$ intersecting $AB$ at $C$... Therefore, for $C$ to NOT lie on any other $PQ$, this equation must have no solution $p, q$ in $(0,1)$ other than the $PQ=AB$ case.
    
    [...]
    
    So, $(-q/p)((1 - t)/2) + q \neq (\sqrt{3}/2)t$... leading to $\sqrt{1 - p^2} [ - (1 - t)/p + 2 ] \neq \sqrt{3} t$.
    Hmm. Alternatively, perhaps express $q$ in terms of $p$ or vice versa. Since $p^2 + q^2 = 1$, so $q = \sqrt{1 - p^2}$.
    
    [...] 
    
    But as I derived earlier, leading to $\sqrt{1 - p^2} [ - (1 - t)/p + 2 ] \neq \sqrt{3} t$. Hmm. Alternatively, perhaps parameterize differently. Wait, hold on. Let me think of $C$ as a point on $AB$...''
    
    \vspace{0.5em}
    \noindent \textit{(The model repeats this derivation loop for over 1,000 tokens before terminating with a random guess.)}
\end{tracebox}

\subsection{\mybasemethod hallucinates contradictions that disrupt inductive reasoning}
This failure mode occurs when the model introduces factual inconsistencies, or ``hallucinations'', into its own reasoning history, directly contradicting valid conclusions it just derived. A single corrupted derivation in the reasoning traces obscures the underlying mathematical structure (e.g., an arithmetic sequence), preventing the model from inducing a general rule. Consequently, it abandons efficient generalization strategies and reverts to inefficient, step-by-step manual simulation that eventually exhausts the context window.

\paragraph{Question:} \textit{Alice and Bob play the following game. A stack of $n$ tokens lies before them. The players take turns with Alice going first. On each turn, the player removes either $1$ token or $4$ tokens from the stack. Whoever removes the last token wins. Find the number of positive integers $n$ less than or equal to $2024$ for which there exists a strategy for Bob that guarantees that Bob will win the game regardless of Alice's play.}

\paragraph{Answer:} 809.

\paragraph{Failure Analysis:} The model correctly computes the status of small $n$ values step-by-step. It correctly derives that $n=21$ is a \textit{winning} position. However, in the subsequent summary step, it hallucinates that $21$ is \textit{losing}. This error breaks the visible pattern (congruence modulo 5), causing the model to doubt its hypothesis and revert to manually simulating the game state for $n > 22$, which is computationally infeasible within the token limit.

\begin{tracebox}
    ``$n=21$: current player can take 1, leaving 20 [which is losing]. So current player can leave opponent in losing—so \textbf{$n=21$ is winning}.
    
    Hmm, so the losing positions up to $n=21$ are:
    0, 2, 5, 7, 10, 12, 15, 17, 20, \textbf{21}...
    
    Looking at differences between consecutive losing positions:
    20-17=3
    21-20=1
    
    So after the initial +2, it's alternating between +3 and +2, except for the last jump which is +1 from 20 to 21. Hmm, that's a bit irregular.
    
    [...]
    
    $n=22$: Current player can take 1, leave 21—which is dp[21]=True... Wait, no—if $n=21$ is a losing position, then opponent would lose...''
\end{tracebox}

\subsection{\mybasemethod fails to recover from a wrong assumption}
Strong guidance can occasionally reinforce ``greedy'' heuristics. The model locks onto a simplified constraint early in the reasoning process. When rigorous algebraic substitution reveals a contradiction, the model exhibits \textit{Algebraic Rigidity}: rather than questioning the heuristic, it assumes it has misread the problem statement, leading to a loop of text re-interpretation.

\paragraph{Question:} \textit{Find the number of ways to place a digit in each cell of a 2x3 grid so that the sum of the two numbers formed by reading left to right is $999$, and the sum of the three numbers formed by reading top to bottom is $99$. The grid below is an example of such an arrangement because $8+991=999$ and $9+9+81=99$.}
\[\begin{array}{|c|c|c|} \hline 0 & 0 & 8 \\ \hline 9 & 9 & 1 \\ \hline \end{array}\]

\paragraph{Answer:} 45.

\paragraph{Failure Analysis:} The model sets up the vertical sum $10(a+b+c) + (d+e+f) = 99$. It applies a greedy heuristic that place values do not overlap, assuming $a+b+c=9$ (tens) and $d+e+f=9$ (units). When substituting values derived from the rows yields a contradiction ($18=9$), the model enters a ``verification loop,'' repeatedly re-calculating the example provided in the prompt rather than revising the incorrect constraint $a+b+c=9$.

\begin{tracebox}
    ``So, $10(a + b + c)$ is 90, and $(d + e + f)$ is 9... So, top-to-bottom constraints:
    1. $a + b + c = 9$
    2. $d + e + f = 9$
    
    [...]
    
    Substitute into constraint 5:
    $(9 - a) + (9 - b) + (9 - c) = 9$
    Simplify: $27 - (a + b + c) = 9$.
    From constraint 4, $a + b + c = 9$, so: $27 - 9 = 18 = 9$? Hmm, that's not matching.
    
    This suggests a contradiction, meaning that my constraints are wrong.
    Wait, maybe I made a wrong assumption... Wait, let me retrace... Wait, hold on, the problem says...
    
    \textit{(The model spends the remaining context re-verifying the example grid: ``Wait, original grid: First row 0, 0, 8...'')}''
\end{tracebox}

\subsection{Mitigations and Future Work}
The failure modes identified above, specifically the tendency to loop or stagnate, suggest that a static guidance strength $\alpha$ may be suboptimal for all reasoning stages. Future work could investigate \textit{dynamic guidance annealing}~\citep{DBLP:conf/nips/FanL0TQCC24}, where $\alpha$ is reduced when repetitive n-grams or circular logic are detected, allowing the target model to ``break out'' of local minima or terminate unpromising paths. Additionally, since hallucination often stems from the model losing track of long contexts, explicitly prompting the guider to summarize intermediate states could reduce hallucination. Finally, for failures in abstraction, equipping the model with tool use (e.g., a Python interpreter) could allow it to offload intractable algebraic verifications that currently lead to context exhaustion.

\section{Impact of the Guidance Strength}
\label{app:hyperparam}
\begin{figure}
  \centering
  \includegraphics[width=0.95\linewidth]{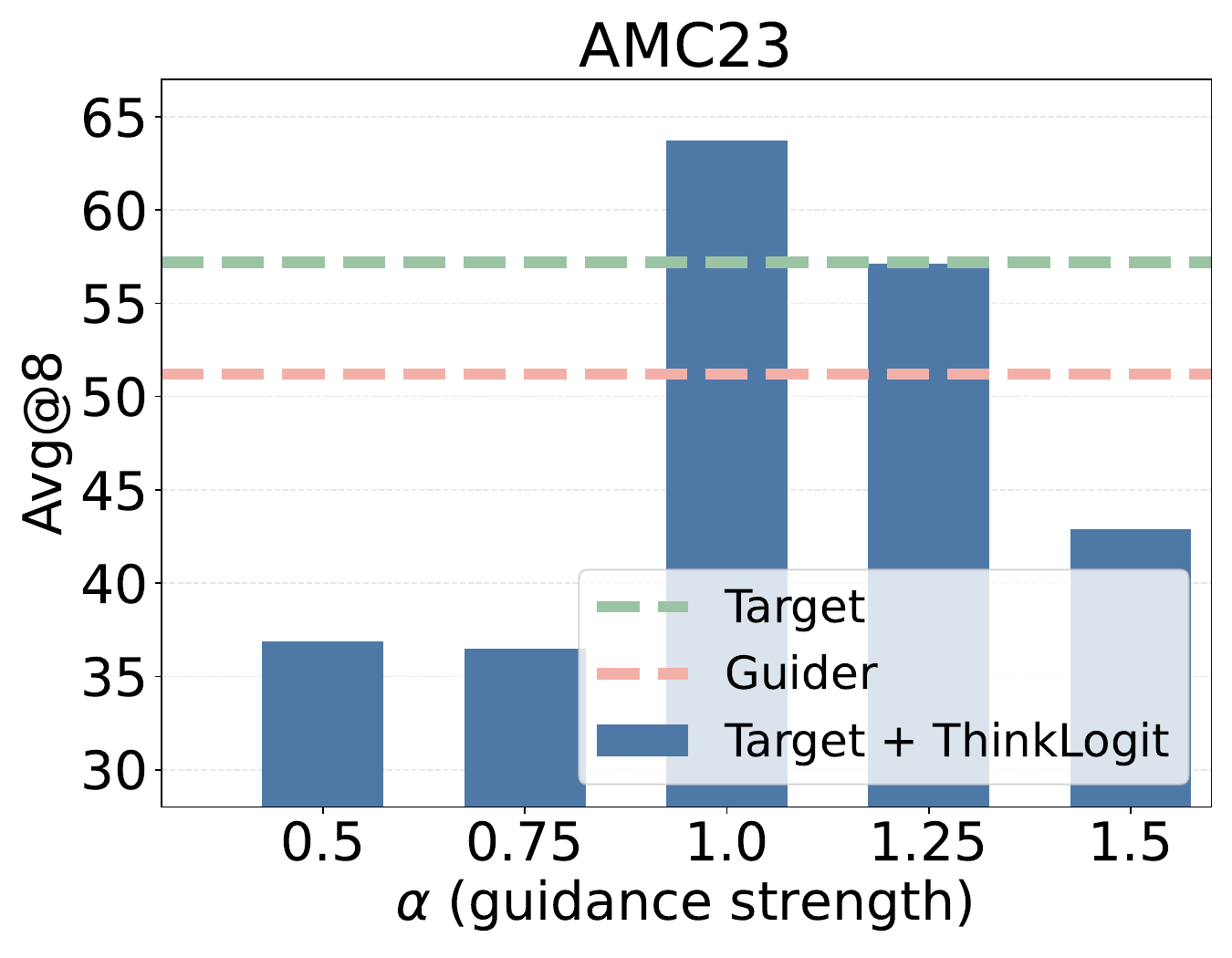}  
\caption{Sweeping the guidance strength $\alpha$ shows that $\alpha$ = 1.0 yields the best trade-off between guider influence and target model priors.}
  \label{fig:alpha}
\end{figure}
In \mybasemethod, the hyperparameter of the guidance strength $\alpha$ plays a critical role. We evaluate its effects on the AMC23 benchmark, which presents a suitable mix of problem difficulties and clearly exhibits both stability and guidance effects.
We sweep $\alpha$ over $\{0.5,0.75,1.0,1.25,1.5\}$ to control how strongly the guider’s delta‑logits modify the target’s distribution (see Figure~\ref{fig:alpha}).  At $\alpha{=}1.0$, we observe the highest avg@8 together with moderate generation length, indicating an optimal trade‑off between the guider’s corrective signal and the target model’s own priors. 
Crucially, all experiments in the main paper except this hyperparameter study use the same hyperparameter $\alpha{=}1.0$ as a robust default, demonstrating that \textit{our method achieves strong performance without extensive hyperparameter tuning}.

\end{document}